\documentclass[journal,twocolumn,a4paper,10pt]{IEEEtran}
\usepackage{cite,url}
\usepackage{mathtools}
\usepackage{amsmath}
\usepackage{amssymb}
\usepackage{amsfonts}
\usepackage{xcolor}
\usepackage{latexsym}
\usepackage{graphicx}
\usepackage{pgfplots}
\pgfplotsset{compat=newest}
\pgfplotsset{plot coordinates/math parser=false}
\newlength\figureheight
\newlength\figurewidth

\definecolor{mid-gray}{gray}{0.5}
\definecolor{green}{RGB}{50,150,0}
\definecolor{blue}{RGB}{0,0,150}

\begin{document}

\newcommand{\etal}{{et al.\;}}

\newcommand{\prob}[1]{{\mathbb{P}\!\left[{#1}\right]}}
\newcommand{\expect}[1]{{{\mathbb{E}}\!\left[{#1}\right]}}
\newcommand{\var}[1]{{{\rm var}\!\left[{#1}\right]}}
\newcommand{\cov}[2]{{{\rm cov}\!\left[{#1},{#2}\right]}}
\newcommand{\cor}[2]{{\rho\!\left({#1},{#2}\right)}}

\newcommand{\eulergamma}{\gamma_{\mathrm{eul}}}
\newcommand{\Ei}[1]{E_i\left(#1\right)}

\newcommand{\R}{\mathbb{R}}
\newcommand{\N}{\mathbb{N}}
\newcommand{\dd}{{\rm d}}
\newcommand{\uniform}{\mathbb{U}(0,1)}

\newcommand{\ppp}{\Phi}
\newcommand{\interf}{i}
\newcommand{\fad}[2]{h_{#2}^2}
\newcommand{\fadsq}[2]{h_{#2}^4}
\newcommand{\ploss}[1]{\ell\left(\left\|#1\right\|\right)}
\newcommand{\plossn}[1]{\ell_{#1}}
\newcommand{\plosssc}[1]{\ell\left(#1\right)}
\newcommand{\plosssq}[1]{\ell^2(\|#1\|)}
\newcommand{\plossxy}{\ell_{xy}}
\newcommand{\plossx}{\ell_{x}}
\newcommand{\plossy}{\ell_{y}}

\newcommand{\channel}[2]{\fad{#1}{#2}\,\plossn{#1}}
\newcommand{\channelx}{\fad{t}{x}\,\plossx}

\newcommand{\tx}[2]{\gamma_{#2}}
\newcommand{\txone}[1]{\gamma^{\rm I}_{#1}}
\newcommand{\txtwo}[1]{\gamma^{\rm II}_{#1}}
\newcommand{\prdensii}[1]{\rho^{(2)}(#1)}
\newcommand{\dens}{\lambda}

\newcommand{\txpwr}{\kappa}
\newcommand{\rxpwr}{p_\mathrm{RX}}
\newcommand{\nakm}{m}
\newcommand{\plc}{\alpha}
\newcommand{\txp}{\mu}
\newcommand{\txpc}{{q}}
\newcommand{\txpd}{\mu}

\newcommand{\chlen}{\eta}
\newcommand{\msglen}{\ell}

\newcommand{\ti}{t}
\newcommand{\delay}{\tau}
\newcommand{\snd}[1]{S(#1)}

\newcommand{\pa}{p_1}
\newcommand{\pab}{p_{12}}
\newcommand{\papa}{p_{1/1}}
\newcommand{\papb}{p_{1/2}}

\newcommand{\areac}{A_c}
\newcommand{\areax}{A_x}
\newcommand{\areay}{A_y}

\newcommand{\cir}{\mathcal{C}}

\newcommand{\expectppp}[1]{{{\mathbb{E}}\!\left[{#1}\right]}}
\newcommand{\expecth}[1]{{{\mathbb{E}}\!\left[{#1}\right]}}
\newcommand{\expecttx}[1]{{{\mathbb{E}}\!\left[{#1}\right]}}

\newcommand{\gap}{g}

\newcommand{\speed}{\bar{v}}
\newcommand{\mobstep}{\omega}
\newcommand{\mob}{{\omega_\delay}}
\newcommand{\expectmob}[1]{{{\mathbb{E}}\!\left[{#1}\right]}}

\newcommand{\cohti}{\delay_c}
\newcommand{\cohth}{\theta}

\title{Interference Prediction in Wireless Networks: \mbox{Stochastic\,Geometry\:meets\:Recursive\,Filtering}}
\author{Jorge F. Schmidt, Udo Schilcher, Mahin K.~Atiq, and Christian Bettstetter,~\IEEEmembership{Senior~Member,~IEEE}
	\IEEEcompsocitemizethanks{
			\IEEEcompsocthanksitem This work was supported with funding from the Austrian Science Fund (FWF) under grant P24480-N15 (Dynamics of interference in wireless networks) and  by  the  K-project  DeSSnet  (Dependable,  secure  and time-aware sensor networks). The latter is  funded  within  the context  of COMET–-Competence Centers  for  Excellent Technologies  by  the  Austrian Federal Ministry for Climate Action, Environment, Energy, Mobility, Innovation  (BMK),  the Federal  Ministry  for  Digital and Economic Affairs (BMDW), and  the  federal  states  of  Styria  and Carinthia; the  COMET program  is  conducted  by  the  Austrian  Research Promotion Agency~(FFG). 
			\IEEEcompsocthanksitem Jorge~F.~Schmidt is with the Institute of Networked and Embedded Systems, University of Klagenfurt, Lakeside Park B02,  and also with Lakeside Labs GmbH, Lakeside Park B04, 9020 Klagenfurt am W\"orthersee, Austria. E-mail: \texttt{Jorge.Schmidt@aau.at}. 
			\IEEEcompsocthanksitem Udo Schilcher and Christian Bettstetter are with the Institute of Networked and Embedded Systems, University of Klagenfurt, Lakeside Park B02, 9020 Klagenfurt am W\"orthersee, Austria.
			\IEEEcompsocthanksitem Mahin K. Atiq is with Silicon Austria Labs GmbH, Austria. Her work has been performed when she was with the Institute of Networked and Embedded Systems, University of Klagenfurt, Austria.
	}
	\vspace{-2em}
}
\maketitle
\thispagestyle{empty}

\begin{abstract}
This article proposes and evaluates a technique to predict the level of interference in wireless networks. We design a recursive predictor that estimates future interference values by filtering measured interference at a given location. The predictor's parameterization is done offline by translating the autocorrelation of interference into an auto\-regressive moving average (ARMA) representation. This ARMA model is inserted into a steady-state Kalman filter enabling nodes to predict with low computational effort. Results show a good accuracy of predicted values versus true values for relevant time horizons. Although the predictor is parameterized for Poisson-distributed nodes, Rayleigh fading, and fixed message lengths, a sensitivity analysis shows that it also tends to work well in more general network scenarios. Numerical examples for underlay device-to-device communications, a common wireless sensor technology, and coexistence scenarios of Wi-Fi and LTE illustrate its broad applicability. The predictor can be applied as part of interference management to improve medium access, scheduling, and radio resource allocation.
\end{abstract}


\section{Introduction}
\label{SEC:INTRO} 

The management of interference has always been a key issue in wireless systems~\cite{Zheng2014}. 
Negative effects of interference are averted by radio resource management, medium access control, scheduling, and decoding techniques. 
Positive aspects may include physical-layer security and energy harvesting from radio waves. In all cases, it seems beneficial to have the ability to {\it predict} interference into the future\,---\,an approach that has not been investigated comprehensively and for which a technique is proposed and evaluated in this article.

Interference can be modeled as a random variable whose properties depend on several parameters, including node locations, mobility, and data traffic. Some properties\,---\,including mean interference, higher-order statistics, and distributions\,---\,can be calculated in a given setup using stochastic geometry~\cite{haenggi11:mean-interf,schilcher16:tit,yang03:tsp,sousa90:jsac,mathar95,mathar1997,net:Ganti09tit}. 
These results consider the spatial features of wireless networks, which makes them fundamentally different from  ``classical" pieces of work on interference modeling and analysis \cite{yao1992investigations,Hamdi2009_SINRstats}. 
A branch of research analyzes how interference {\it changes\/} over time and space~\cite{ganti09:interf-correl,schilcher12:interfcor,haenggi13:div-poly,tanbourgi14:mrc,tanbourgi15:phd,koufos2018:tmc,schilcher18:cohertime}. Such interference dynamics can be expressed in terms of the autocorrelation of the received interference. Correlation influences the system behavior, such as the performance of diversity, relaying, multiple-input multiple-output (MIMO), and medium access protocols (see \cite{haenggi2012diversity,haenggi13:div-poly,tanbourgi14:mrc,tanbourgi14:nakagami}).  

Despite advances in the modeling of interference dynamics, this knowledge has not been exploited to actually improve the performance of wireless systems~\cite{atiq17:mswim}. The state of research is not as advanced as in channel modeling, where insights on channel dynamics\,---\,such as coherence time and decorrelation distances\,---\,are indeed used in practice (e.g., space-time coding and MIMO). 
Taking this step from modeling to design is the goal of our research: the investigation of  {interference prediction}. The fundamental question at the core of our work is: ``How well can one predict, in a probabilistic manner, the interference power at a given location in a certain network?'' Initial steps in this direction were made in \cite{atiq17:mswim}, which proposes a simple prediction technique based on learning of traffic patterns, and by others in \cite{7076612}, which proposes prediction  based on the mobility of nodes without considering traffic and channel. Furthermore, contention-based medium access control protocols perform some type of \textit{implicit} estimation of future interference levels through their inherent (exponential) back\-off techniques. 

\begin{figure}[t]
 \centering
  \includegraphics[width=.48\textwidth,keepaspectratio=true]{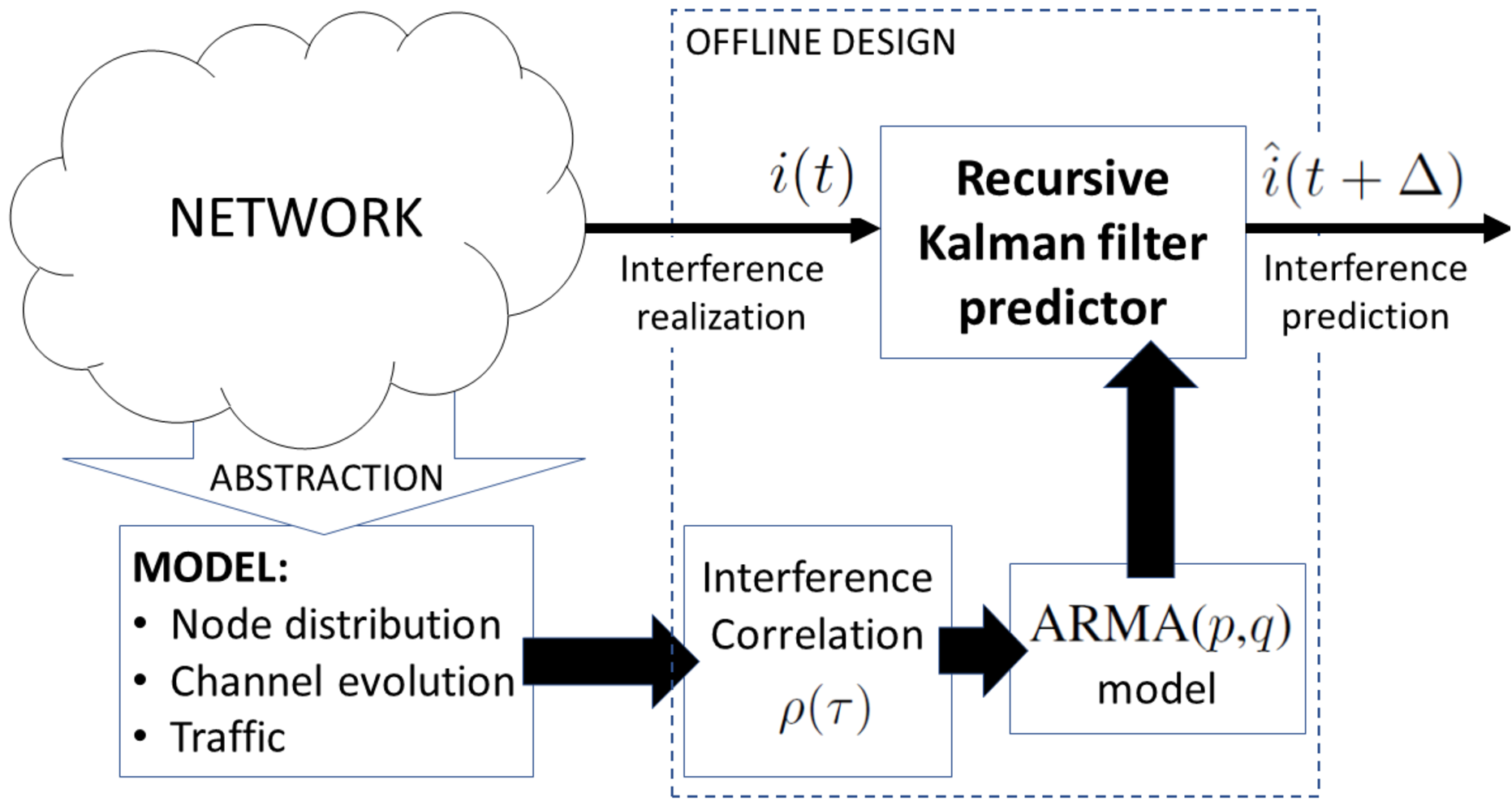}
\caption{Block diagram of the interference predictor.} 
\label{Fig:BlockDiagram}
\end{figure}

Our approach for predicting the level of interference is to  merge results on the autocorrelation of interference~\cite{schilcher18:cohertime} with recursive filtering. The specific contributions are as follows:
\begin{itemize}
\item A method is presented to map the autocorrelation function of interference into an auto\-regressive moving average (ARMA) model suited for performing forecasts from previous interference observations. This mapping is calculated for Poisson distributed nodes, Rayleigh fading, and random medium access with fixed message lengths.  \item An offline and blind predictor is obtained by inserting the ARMA representation of the interference process into a Kalman filter. The time invariance of the model leads to a steady-state gain of the Kalman predictor to perform lightweight predictions at individual nodes. A block diagram of the predictor design is given in Fig.~\ref{Fig:BlockDiagram}.
\item Simulations show that this predictor outperforms both basic predictors and predictors that consider only the channel dynamics (and disregard the impact of~traffic).  
\item A sensitivity analysis demonstrates the robustness of the predictor against certain inaccuracies and model mismatches in its parameterization: it performs well with traffic and node distribution models that are more general than the ones used in its design.
Numerical examples for underlay device-to-device communications, a wireless sensor network, and coexistence scenarios of Wi-Fi and LTE  illustrate the applicability of the proposed  scheme.
\end{itemize}

The rest of the article is organized as follows. Section~\ref{SEC:MODEL} introduces the system model. Section~\ref{SEC:Predictor} derives the interference predictor, starting with the stochastic geometry models of interference and progressing to the low-complexity recursive predictor implementation. Section~\ref{SEC:Evaluation} evaluates the predictor  in terms of accuracy and includes a sensitivity analysis and specific technology examples. Finally, Section~\ref{SEC:Conclusions} concludes.

\section{System Model}
\label{SEC:MODEL} 

The system is described by the placement and mobility of nodes, their data traffic behavior, and the radio channel. Time is discretized into slots $\ti \in \mathbb{N}_0$. The parameterization of the predictor is done for the following  \textit{base system}.

Node positions follow a two-dimensional Poisson point process (PPP) $\ppp=\ppp(t)$ of intensity~$\lambda$. All nodes move with speed~$\nu$ following a time-discrete Brownian motion. 
This mobility model is chosen because it preserves the uniform spatial distribution: the node locations at a given time $t$ form a PPP $\ppp(t)$ with intensity $\lambda$~\cite{gong14:tmc}. Other models with this property could also be applied. The location $x$ of a node at time $t+\tau$ is $x_{t+\tau}=x_t + \nu\omega_\tau$,~with $\omega_\tau = \sum_{t=1}^\tau \omega_t \stackrel{d}{=} \sqrt{\tau}\omega_0$, where the last equal sign denotes equality in distribution and the $\omega_t$ are i.i.d.~two-dimensional Gaussian random variables of zero mean and covariance~matrix 
\begin{equation}
	\Sigma = \left(
	\begin{array}{ccc}
	0  & \sqrt{{2}/{\pi}}  \\
	\sqrt{{2}/{\pi}} & 0 
	\end{array}
	\right).
\end{equation}

At a given slot $t$, each node is either idle or transmitting with power $\txpwr$ such that, on average, a fraction $\mu$ of all nodes starts a new transmission. The message duration $\msglen$ is the same for all transmissions. In any slot, each idle node starts a transmission with probability $\frac{\txp}{1-\txp(1-\msglen)}$. This yields an expected traffic intensity of $\txp\msglen$ interferers in each slot~\cite{schilcher18:cohertime}. 
Such a combination of node placement and random access is often used to model wireless sensor networks and (with some considerations) cellular networks, in particular for device-to-device communications (see \cite{Lin2013, ElSawy2014, Lee2015, Erturk2013, Ye2014, schmidt2015underlay}).

The channel is described by a standard model with distance-dependent attenuation and small-scale fading due to multipath propagation. 
The path gain is  $g(x,\xi)=\min(1,\left\|x-\xi\right\|^{-\alpha})$ with path loss exponent $\alpha > 2$ and some normalization. Small-scale fading is modeled by Rayleigh fading with $\expect{h^2(\ti)}=1$. The channel gain $h^2(\ti)$ is then exponentially distributed. The Jakes-Doppler model~\cite{Jakes} describes the continuous time evolution of the wireless channel. It assumes uniform scatterers' directions which makes it fit well for a wide range of propagation environments. Its amplitude's autocorrelation function~is $\rho_J(\tau)\!=\!J_0(2\pi\tau\nu_{\rm max})$, where $J_0(\cdot)$ is the zeroth-order Bessel function of the first kind and $\nu_{\rm max}=\nu$ is the maximum expected speed. The channel coherence time $\chlen$ is the time lag for which the channel autocorrelation first reaches a small threshold, which is zero in this~article. 

A message transmitted at location $x$ causes interference at $\xi$ with power $\interf_{\xi_x}(\ti) = \txpwr \: g(x,\xi) \: h_x^2(\ti)$. 
The overall interference at location $\xi$ at time $t$ is the sum of the reception powers from all transmitting nodes. In Poisson networks, the consideration of a typical location~$\xi$ is equal to the consideration at the origin of the plane $\R^2$ due to Slivnyak's theorem  \cite{stoyan95:_stoch_geomet_applic, haenggi09:interference}. In the given base system, the overall interference power is thus
\begin{equation}
\interf(\ti)=\sum_{x\in\ppp}i_{\xi_x}(\ti)\gamma_x(\ti) =\sum_{x\in\ppp}\txpwr\,\min(1,\|x\|^{-\alpha})\,h_x^2(\ti)\,\gamma_x(\ti)\,,
\label{eq:interf_def}
\end{equation}
where $\gamma(\ti)$ is a Bernoulli random variable indicating whether node $x$ is sending ($\gamma_x(\ti)=1$) or not ($\gamma_x(\ti)=0$), condensing the information from the parameters $\txp$ and $\msglen$. The interference process in \eqref{eq:interf_def} is stationary \cite{haenggi09:interference} and changes over time due to the time-varying characteristics of node locations, wireless channel, and traffic. These sources of time-varying behavior are captured by the parameters $\nu$, $\chlen$, $\txp$, and~$\msglen$.

\section{Interference Prediction}
\label{SEC:Predictor} 
\subsection{Interference Correlation}

A main approach for channel prediction is based on  exploiting the autocorrelation  of the channel~\cite{Hallen_Dec07}. We take a similar approach for interference prediction: our design starts with an analytical model for the autocorrelation  associated to the interference process. We use Pearson's correlation coefficient of $\interf(\ti)$ for two time instants $\ti_1$ and $\ti_2$, which  is:
\begin{equation}
	\label{eq:PearsonCoeff}
\cor{\interf({\ti_1})}{\interf({\ti_2})}=\frac{\cov{\interf({\ti_1})}{\interf({\ti_2})}}{\sigma_{\interf}^2}\:,
\end{equation}
where $\cov{\interf({\ti_1})}{\interf({\ti_2})}=\expect{\interf({\ti_1})\interf({\ti_2})}-\expect{\interf({\ti_1})}\,\expect{\interf({\ti_1})}$ is the covariance of $\interf$, $\sigma_{\interf}^2$ its  variance, and $\expect{i}$ the expected value. The  lag is denoted by $\delay=\ti_2-\ti_1$.

Expressions for this correlation are known for different system models~\cite{schilcher18:cohertime}. The model that parameterizes our predictor corresponds to Case~$(2,2,2)$ in \cite{schilcher18:cohertime}, for which 
\begin{equation}
	\label{eq:AutocorrFunction}
	\begin{split}
	\cor{\interf(\ti_1)}{\interf(\ti_2)} 
	&=\frac{\left(J_0^2(2\pi\tau\nu_{\rm max})+1\right)\expect{\gamma(\ti_1)\,\gamma(\ti_2)}}{2\txp\msglen}\:\cdot\\
	&\quad \frac{(\alpha-1)\int_{\mathbb{R}^2}g(x)\expect{g(x+\nu\omega_\tau)}{\rm d}x}{\alpha\pi},
	\end{split}
\end{equation}
where stationarity implies $\cor{\interf(\ti_1)}{\interf(\ti_2)}=\rho(\interf(\ti_2-\ti_1))$, which we denote $\rho(\tau)$. The integral in the last line of the equation can only be solved numerically in case of mobility.
The traffic contribution is~\cite{schilcher18:cohertime}
\begin{eqnarray}\nonumber
\lefteqn{\expect{\gamma(\ti_1)\,\gamma(\ti_2)}\!=\!\max\big(0,\txp\,(\msglen-\delay)\big)\!+\!\frac{\txp^2}{1\!-\!\txp(\msglen\!-\!1)}}\\\nonumber
&&\!\!\!\!\!\!\!\!\!\!\!\sum_{i=0}^{\min(\delay-1,\msglen-1)}  \sum_{j=1}^{\min(\delay-i,\msglen)} \sum_{k=0}^{\left\lfloor\frac{\gap}{\msglen}\right\rfloor}\!\! \binom{\gap\!-\!k\msglen\!+\!k}{k}\beta^{\gap-k\msglen}(1\!-\!\beta)^k\:, \nonumber \end{eqnarray}
with $\gap=\delay-i-j$, and $\beta=1-\txp/(1-\txp(\msglen-1))$ being the probability of a node staying idle in a slot.
From~\eqref{eq:AutocorrFunction} and  \cite[Th.2]{schilcher18:cohertime}, the interference correlation when the channel is the sole source of correlation yields $\rho(\tau)=J_0^2(2\pi\tau\nu_{\rm max})$. 

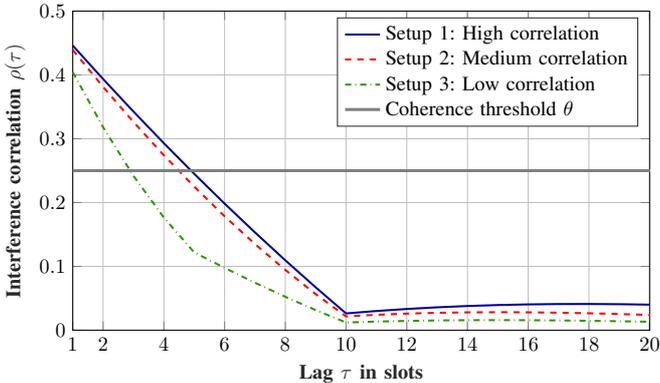
\begin{figure}[htbp]
 \centering
  \resizebox{0.49\textwidth}{!}{
\begin{tikzpicture}
	\begin{axis}[%
	width=4.527in,
	height=2.8in,
	xmin=1,
	xmax=20,
	xtick={ 1,  2,  4,  6,  8, 10, 12, 14, 16, 18, 20},
	xlabel style={font=\bfseries\color{white!15!black}},
	xlabel={Lag $\tau$ in slots},
	ymin=0,
	ymax=0.5,
	ylabel style={font=\bfseries\color{white!15!black}},
	ylabel={Interference correlation $\rho(\tau)$},
	axis background/.style={fill=white},
	xmajorgrids,
	ymajorgrids,
	yticklabel style={/pgf/number format/fixed},
	legend style={legend cell align=left, align=left, draw=white!15!black}
	]
	\addplot [color=blue, line width=1.0pt]
		table[row sep=crcr]{%
	1	0.44603082102369\\
	2	0.393563397628\\
	3	0.342584772993419\\
	4	0.293081547579476\\
	5	0.245039912120347\\
	6	0.198445680899148\\
	7	0.153284314335661\\
	8	0.109540953700069\\
	9	0.067200440609227\\
	10	0.0262473454144541\\
	11	0.0300476985927673\\
	12	0.0332342048439109\\
	13	0.0358358315591083\\
	14	0.0378809858070639\\
	15	0.0393975091121739\\
	16	0.04041267512887\\
	17	0.0409531875795892\\
	18	0.0410451796403066\\
	19	0.0407142130980667\\
	21	0.0393024972655063\\
	};
	\addlegendentry{Setup 1: High correlation}

	\addplot [color=red, dashed, line width=1.0pt]
		table[row sep=crcr]{%
	1	0.439206867457493\\
	2	0.381290555686881\\
	3	0.326265632129363\\
	4	0.274137688673736\\
	5	0.224904200433933\\
	6	0.178555310850783\\
	7	0.13507453597072\\
	8	0.0944394090127219\\
	9	0.0566220698810049\\
	10	0.021589799485227\\
	11	0.0240996382688792\\
	12	0.0259583735115072\\
	13	0.0272223603794011\\
	14	0.0279467240775517\\
	15	0.0281852513859242\\
	16	0.0279902972332948\\
	17	0.0274127023057176\\
	18	0.0265017328450199\\
	19	0.0253050182232677\\
	20	0.02386850803207\\
	21	0.023659426317046\\
	};
	\addlegendentry{Setup 2: Medium correlation}

	\addplot [color=green, dashdotted, line width=1.0pt]
		table[row sep=crcr]{%
	1	0.404380449475884\\
	2	0.317986848973309\\
	3	0.241846132222385\\
	4	0.1764634737211\\
	5	0.121956783720005\\
	6	0.0976928589878092\\
	7	0.0744445482316749\\
	8	0.0523548598527093\\
	9	0.0315339991432424\\
	10	0.0120651882124569\\
	11	0.0135001393894463\\
	12	0.0145625539619303\\
	13	0.0152806860668839\\
	14	0.0156841306417483\\
	15	0.0158033094389971\\
	16	0.0156690420861807\\
	17	0.0153121840932897\\
	19	0.0140524784653309\\
	21	0.0123932814666006\\
	};
	\addlegendentry{Setup 3: Low correlation}

	\addplot [color=mid-gray, solid, line width=1.5pt]
		table[row sep=crcr]{%
	1	0.25\\
	21	0.25\\
	};
	\addlegendentry{Coherence threshold $\theta$}

	\end{axis}
\end{tikzpicture}%
}
\caption{Interference correlation for the setups of Table~\ref{tab:Scenarios}. For a threshold $\theta = 0.25$, the interference coherence time is $\tau=5$ slots (Setups 1 and 2)  and $\tau=3$~slots (Setup~3).} 
\label{Fig:Interf_correl}
\end{figure}

Similar to the characterization of  channels, one can define the interference coherence time $\tau_c$ to be the lag until
the correlation of interference is smaller than a threshold~$\theta$~\cite{schilcher18:cohertime}, i.e., $\tau_c = {\rm min}\{\tau \in \N \ | \ \rho(\tau)\leq \theta\}$. In contrast to the channel coherence time, there are scenarios for which the interference correlation does not reach zero~\cite{schilcher18:cohertime}. To account for this, we set $\theta = 0.25$ as the threshold, which is reached by all scenarios in this article.
Table~\ref{tab:Scenarios} shows three combinations of parameter values to be used to yield a high, a moderate, and a low correlation. The interference correlation for these setups is shown in Fig.~\ref{Fig:Interf_correl}. These setups are used to illustrate some design steps and assess the performance of the~predictor. 

\begin{table}[th]
\begin{center}
\renewcommand{\arraystretch}{1.3}
\caption{System Model Parameters}
\label{tab:Scenarios}
\begin{tabular}{ccccc}
\hline
{Setup} & Speed  & Traffic pa- & Message & Channel coher- \\
 & $\nu$ & rameter $\txp$ & length $\msglen$ & ence time $\chlen$ \\\hline
{1} & 0.77 cm/slot& 0.01 / slot & 10 slots & 50 slots \\ 
{2} & 1.91 cm/slot& 0.01 / slot & 10 slots & 20 slots \\ 
{3} & 7.65 cm/slot& 0.01 / slot & 10 slots & ~5 slots \\ \hline
\end{tabular}
\end{center}
\end{table}

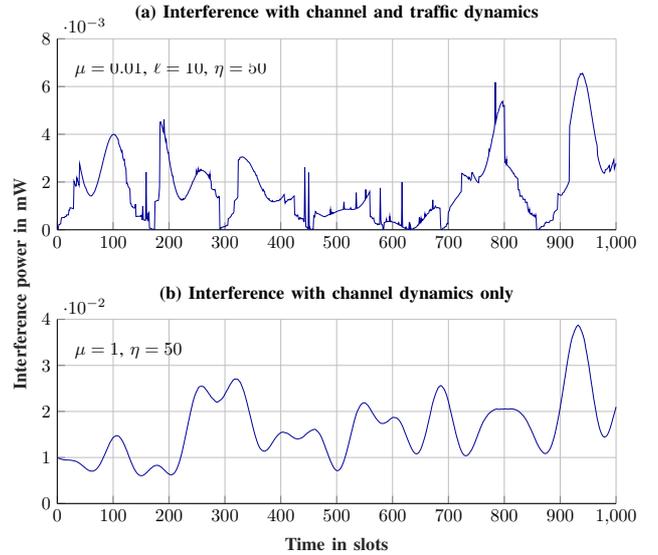
\begin{figure}[t]
 \centering
  \resizebox{0.49\textwidth}{!}{
  \begin{tikzpicture}

\begin{axis}[%
width=4.875in,
height=2.088in,
at={(0.758in,3.2in)},
xmin=0,
xmax=1000,
ymin=0,
ymax=0.008,
axis background/.style={fill=white},
title style={font=\bfseries},
title={(a) Interference with channel and traffic dynamics},
axis x line*=bottom,
axis y line*=left,
xmajorgrids,
ymajorgrids
]
\addplot [color=blue, forget plot]
  table[row sep=crcr]{%
1	0\\
2	8.2947887563023e-05\\
3	0.000222421042622045\\
4	0.000224347439711892\\
5	0.000235617371345143\\
6	0.000250065280852141\\
7	0.000440629647414426\\
8	0.000510451415834723\\
10	0.000508502526940902\\
11	0.000513594707058473\\
13	0.000511343050902724\\
14	0.000560383551032828\\
16	0.000559989359089741\\
17	0.000627622229217195\\
20	0.000620489213929432\\
21	0.000822846244091124\\
22	0.00082130613566278\\
23	0.000924487552993014\\
26	0.000905170059354532\\
29	0.00088120132545555\\
30	0.00202906612219067\\
31	0.00195391214197116\\
32	0.00189736922254724\\
33	0.00182552447995477\\
34	0.00183304442305143\\
35	0.00218160166627968\\
36	0.00210232458869086\\
37	0.00201947086975451\\
39	0.00186370329549845\\
40	0.0027602722454958\\
41	0.00264321605868645\\
43	0.00242030087929379\\
44	0.00231249795388067\\
46	0.00210610464637284\\
47	0.00200995026216333\\
48	0.00191957131607978\\
49	0.00186045591135553\\
50	0.00178538683690022\\
51	0.00176121849801802\\
52	0.00169133389283616\\
53	0.00162861772821543\\
54	0.00157431675143016\\
57	0.00146550215310981\\
58	0.0014404387804916\\
59	0.00142415175116639\\
60	0.00141529975019239\\
61	0.00141463874319925\\
62	0.00143481152088043\\
63	0.00145872443818007\\
64	0.00149759243799963\\
65	0.00152934996447129\\
66	0.00156720550671707\\
67	0.00161214148761246\\
68	0.00166273531681327\\
69	0.00173900633046742\\
70	0.00180295895711424\\
71	0.00187346533368782\\
72	0.00194753007940562\\
73	0.00203175933643251\\
75	0.0021954614093147\\
76	0.00228421711460669\\
77	0.00239679564253947\\
78	0.00247182711723326\\
82	0.00286185908521475\\
83	0.00296033798952067\\
87	0.00332455015495725\\
89	0.00348752940374197\\
90	0.00356481601102132\\
91	0.00363627006095157\\
92	0.0037000583089366\\
93	0.00376061229712832\\
94	0.0038136787171652\\
95	0.00386049080418616\\
97	0.00393635312445895\\
98	0.00396721333811456\\
100	0.0039918969844166\\
101	0.00399719077597638\\
102	0.00399401639549524\\
103	0.00396878845685933\\
104	0.00396886416660891\\
105	0.00393104916258835\\
106	0.00389948555277897\\
107	0.00386282505883173\\
108	0.0038031253209283\\
109	0.00378876085608226\\
110	0.00360290035689559\\
111	0.00355713531075708\\
112	0.00348853491721002\\
113	0.00343619502530146\\
114	0.00334433728733075\\
115	0.00326334031660735\\
116	0.00331012620915772\\
117	0.00309918633513462\\
118	0.0029768065829785\\
119	0.0029203137529521\\
120	0.00294678496948109\\
121	0.00270803545708986\\
122	0.00261581674419631\\
123	0.00253432034514844\\
124	0.00193700280738085\\
125	0.00195831082669429\\
126	0.00189874392935963\\
127	0.00181313561574825\\
128	0.00177384974540473\\
130	0.0017046887143124\\
131	0.00110260467829448\\
132	0.00109254634446643\\
133	0.00099725534062145\\
134	0.000966152783576035\\
135	0.000969594043226607\\
136	0.000927092583538069\\
138	0.000907506431531147\\
139	0.000915067534037917\\
140	0.00107584391719229\\
141	0.000554871464942153\\
143	0.000558453970256778\\
144	0.00057159746211255\\
145	0.000564288788041267\\
147	0.000564558301107354\\
148	0.000566004829124722\\
149	0.000589227296131867\\
150	0.000559760873670712\\
151	0.000592595217199232\\
152	0.000420581824073452\\
153	0.00099388232547426\\
154	0.000437322708876309\\
155	0.000407703535302062\\
156	0.000400743341742782\\
158	0.000393363187754403\\
159	0.0024153861194236\\
160	0.000442533956174884\\
161	0.00070483284605416\\
162	0.000377049780126981\\
163	0.000387181163205241\\
164	0.000754998926481676\\
165	7.42501511012961e-05\\
166	6.54832246027581e-05\\
167	6.46883618173888e-05\\
168	4.20099021312126e-05\\
169	4.31169576131651e-05\\
170	2.11350575227698e-05\\
174	2.17537699427339e-05\\
175	0.0011837914220223\\
177	0.00120419642576053\\
180	0.00122461746798308\\
181	0.00123037051696429\\
182	0.00144755693747811\\
183	0.00175972877116237\\
184	0.00450873290549225\\
185	0.00452164877629002\\
186	0.00444904292146475\\
187	0.00436496470285874\\
189	0.00418218123309089\\
190	0.00410314899011155\\
191	0.00462411339867685\\
192	0.00404520805852826\\
193	0.00385415298012504\\
194	0.0036994512063302\\
195	0.00377854586395188\\
196	0.00346515372291378\\
197	0.00335884828996313\\
198	0.00322144765289067\\
199	0.00323487581044901\\
200	0.0029873169860366\\
201	0.00286558430627792\\
202	0.00273263828159998\\
203	0.00259090596068745\\
204	0.00247101654747439\\
205	0.00235749720036438\\
206	0.00225204980370108\\
207	0.00224899704892323\\
208	0.00203589521788672\\
210	0.00184596951976346\\
211	0.00175767118366821\\
212	0.0016788285572602\\
214	0.00153813714598527\\
215	0.00147664495455047\\
216	0.00142446322706746\\
217	0.00137842981007452\\
218	0.00136369403458048\\
219	0.00130639026576773\\
220	0.00129508105510467\\
221	0.00125696915984008\\
222	0.00128394710236535\\
223	0.00124219183646801\\
224	0.00125751930477236\\
225	0.00129967519785623\\
226	0.00126025078156999\\
227	0.00129472865489788\\
228	0.00130032602373831\\
229	0.00133076304257429\\
230	0.00139259557909099\\
231	0.00142355720527121\\
232	0.00146475798101164\\
233	0.00149042988493875\\
234	0.00153763313755917\\
235	0.00160753953105086\\
236	0.00163927953133225\\
238	0.00174658306139008\\
239	0.00180471589658282\\
241	0.00191382075115598\\
242	0.00200360157475643\\
243	0.00202097343321839\\
246	0.00217075406646927\\
247	0.00222617159784022\\
248	0.00226774425516396\\
249	0.00246958885566073\\
250	0.00233732492495164\\
251	0.00238263780511261\\
252	0.00255541431545225\\
253	0.00243607276263447\\
254	0.00244422170123926\\
255	0.00250147308076976\\
256	0.00247162686218871\\
257	0.00252677159903669\\
258	0.00248615927716855\\
259	0.00248167035567803\\
260	0.00249026640392458\\
261	0.00247165840744401\\
262	0.00246218082475025\\
263	0.00244566178935202\\
264	0.00243485070325278\\
265	0.00240152230014701\\
266	0.00237722291547016\\
267	0.00237147022585305\\
268	0.00232264738588128\\
269	0.00229193845370901\\
270	0.00230028082478384\\
271	0.00219733324297522\\
272	0.00216195238556338\\
273	0.00234258086857153\\
274	0.00209156321807313\\
275	0.00205570763205287\\
276	0.00172927739288298\\
278	0.00168737841693201\\
279	0.00168167507183625\\
280	0.00168215764904289\\
281	0.00166963772915096\\
282	0.00168673211260284\\
283	0.00161203936761467\\
284	0.00144818190722162\\
285	0.00138745239496529\\
286	0.0013738697783765\\
287	0.00133169868263394\\
288	0.00134691554774236\\
289	0.00140040152496113\\
290	0.00138499175977813\\
291	8.36933327263978e-06\\
292	4.79399159303284e-06\\
293	9.16688052257086e-05\\
294	8.79075039392774e-05\\
295	0.000111067586999525\\
296	0.000109400335531973\\
297	0.000114513509743119\\
298	0.00013431086563287\\
300	0.00012571601939726\\
301	0.000160337223178431\\
302	0.000157243106059468\\
303	0.000182086969175543\\
305	0.000186201609380987\\
306	0.000466018931092549\\
308	0.000495218568403288\\
309	0.000511209732508178\\
310	0.000547010214745569\\
311	0.000564712317668636\\
312	0.000633006621683307\\
313	0.000650653888669694\\
314	0.000673317107043658\\
315	0.000690381366780457\\
316	0.000752452611209264\\
317	0.00119560518339767\\
318	0.00122714694964543\\
320	0.00127112096004112\\
321	0.00130000203364489\\
322	0.00132446895395333\\
323	0.00136561064903162\\
324	0.00292475693311189\\
325	0.00294048833427496\\
326	0.00294251909656396\\
327	0.00303206929311273\\
329	0.00303649918691917\\
330	0.00303584127539125\\
331	0.00305473569278547\\
333	0.00303891151611424\\
334	0.00302730749899638\\
336	0.00299240244862631\\
337	0.00296888175068943\\
338	0.00294074032501612\\
339	0.00292948647859248\\
340	0.00289822591753364\\
341	0.00286182645970712\\
342	0.00285651977094403\\
343	0.00282550348299537\\
344	0.00280496845289235\\
345	0.00276447252826983\\
348	0.00263213472044299\\
349	0.00258300202710871\\
350	0.00252961144030905\\
351	0.00248876995965475\\
352	0.00243575645833971\\
353	0.00243971739951121\\
356	0.00226543603287155\\
357	0.00227777408190377\\
358	0.00215844867670967\\
359	0.00211629804664426\\
361	0.00198244093553512\\
364	0.00181998657990334\\
365	0.00176973586223994\\
366	0.00174016290156942\\
367	0.00166940547808281\\
368	0.00163151306128384\\
369	0.00167681923812779\\
370	0.00160851404689311\\
371	0.00160381820990096\\
372	0.00145635940771172\\
373	0.00145386593680996\\
374	0.00138560175219027\\
376	0.00132535598913819\\
377	0.00129891329038401\\
378	0.00128197565265964\\
379	0.0013021964687141\\
380	0.00123377228771915\\
382	0.00120317660378078\\
384	0.00118307303318943\\
385	0.00122127559404817\\
386	0.00117359944010786\\
388	0.00117385292890049\\
389	0.00117793524907484\\
392	0.00120307239797057\\
393	0.00122501180078416\\
394	0.00123206729676895\\
395	0.00120817022002484\\
397	0.00121985974169547\\
398	0.00127460874693952\\
399	0.00125625723546818\\
400	0.00128390887186924\\
401	0.0013160902145728\\
402	0.00133864377653481\\
405	0.00143655008866972\\
406	0.00147149966858251\\
407	0.00124552332670191\\
408	0.00111612403100025\\
409	0.00119377704083945\\
410	0.00118385926782594\\
411	0.00118804389933302\\
412	0.00122897544360967\\
413	0.00126570169061324\\
414	0.00126835879927967\\
415	0.00124309753846319\\
416	0.00128049576073863\\
417	0.00129245251093835\\
418	0.00126099162616811\\
419	0.00128514639254718\\
421	0.00135163770642066\\
422	0.00137226866843321\\
423	0.00139803197896526\\
424	0.00141632847055462\\
425	0.000621355593693806\\
426	0.000625009197278814\\
427	0.000621214252078062\\
428	0.000567337618235797\\
429	0.000573037086724071\\
430	0.000621050269387524\\
431	0.000584947692914284\\
432	0.000441765873461009\\
433	0.000437021029028983\\
434	0.000506467150671597\\
435	0.000510139658558728\\
436	0.000460691516991574\\
437	0.00044959252636545\\
438	0.000477806368849087\\
439	0.00046480033017815\\
440	0.0004614426459284\\
441	0.000340158293056447\\
442	0.000343649435080806\\
443	0.00262465338187212\\
444	0.000656572780485476\\
445	0.000120354166710968\\
446	0.000118682252832514\\
447	9.34133908003787e-05\\
448	5.41091875447819e-05\\
449	7.46215280287288e-05\\
450	0.00239081293193522\\
451	5.24284345146953e-05\\
452	0.000110774127620061\\
453	3.70552161257365e-05\\
454	1.38778635800918e-05\\
455	1.1378021326891e-05\\
457	2.76191492503131e-05\\
458	1.36815323230621e-05\\
459	0.000585966653602554\\
460	0.000563953368100556\\
461	0.000537325112077269\\
462	0.000764767642863262\\
463	0.000739738256811506\\
465	0.000671308554160532\\
466	0.000677474508734122\\
467	0.000670468209364117\\
468	0.000658665314290374\\
469	0.000654797803463225\\
470	0.000637061653378623\\
471	0.000614752675119234\\
472	0.000618028413214233\\
474	0.000585559866181029\\
476	0.000560078715693635\\
478	0.00055209439437931\\
479	0.000545330471481975\\
480	0.000561243622087204\\
481	0.00056306631768166\\
482	0.000591111473568162\\
483	0.000591250975389812\\
484	0.000615548193195536\\
485	0.000617499575355396\\
486	0.000630515979310076\\
487	0.000639629362012784\\
488	0.000632485551591344\\
489	0.000850135818836861\\
490	0.000649679335879227\\
491	0.000669362158987497\\
492	0.000728008565488381\\
493	0.000679098387877275\\
494	0.000719994807354851\\
495	0.000700468931540854\\
498	0.000735954384822435\\
499	0.000761441137910879\\
500	0.000759535643396703\\
502	0.000780921320028938\\
506	0.000814871386637606\\
510	0.000839626506603963\\
511	0.000844304344809643\\
512	0.000862616665358473\\
513	0.0011671724006419\\
514	0.000855913976579359\\
516	0.000863488764935028\\
517	0.00085743438830832\\
518	0.000858372718425926\\
519	0.000856122468235299\\
521	0.000859542286093529\\
522	0.000851565030075108\\
523	0.000856735783713702\\
524	0.000865756134089679\\
525	0.000918845125397638\\
526	0.000848740062338038\\
527	0.000854559448725922\\
528	0.000846955523229553\\
529	0.000848718439101503\\
530	0.000857893919487651\\
531	0.000853155946401785\\
533	0.00086401763462618\\
534	0.00088099781623896\\
535	0.00155286177414382\\
536	0.000900708508652315\\
537	0.000898504230349317\\
539	0.000928140208770856\\
541	0.000963106365361455\\
543	0.00100750214926393\\
544	0.00103573953253999\\
545	0.00105803242399816\\
546	0.0010860831955597\\
547	0.00112124200313701\\
548	0.00114786449955773\\
549	0.00120025838623405\\
550	0.00121551160236777\\
551	0.00125312419277179\\
552	0.00168586922779923\\
553	0.00133075498877133\\
554	0.00137078207183094\\
555	0.00142790563040762\\
556	0.00145945880865384\\
557	0.00150260278610403\\
558	0.00152807205029148\\
559	0.00158766149877465\\
560	0.000542251198908161\\
561	0.000543687453500752\\
562	0.000554627609972158\\
563	0.00051443166955778\\
564	0.00046565611899041\\
565	0.000466635438101548\\
566	0.000492869397362483\\
567	0.000384237674438737\\
568	0.000311605395609149\\
569	0.000271506022841095\\
570	0.000252721717060922\\
572	0.000258587700614044\\
573	0.000209588222674029\\
574	0.00021490142046332\\
575	0.000249438113769429\\
576	0.000224401375930938\\
577	0.000228795803081994\\
578	0.00175050180155267\\
579	0.000402793929993095\\
580	0.000234184236546753\\
581	0.00020690029157322\\
582	0.000206672679382791\\
583	4.81281449538074e-06\\
584	4.94062487632618e-06\\
585	0.000299476657232844\\
586	0.000306735991557616\\
587	0.000318684747185216\\
588	0.00032532090915538\\
589	0.000449414448098651\\
590	0.000334865764216374\\
591	0.000338961935199222\\
592	0.000347359205761677\\
594	0.000345117746746837\\
595	0.000568624539482698\\
596	0.000343521265676827\\
597	0.000368128972354498\\
598	0.000337593280391957\\
601	0.000321078160709476\\
603	0.000305521348536786\\
604	0.00031890389311684\\
605	0.000286203292716891\\
608	0.000251901530305076\\
609	0.000250695431418535\\
610	0.000226125066888017\\
611	0.000297526690701488\\
612	0.000199159915268865\\
613	0.000192746836546576\\
614	0.000178735865461022\\
615	0.000212526906693711\\
616	0.000163281207619548\\
617	0.00199706139380851\\
618	0.000114833114025714\\
619	0.000110642616050427\\
620	8.91564429821301e-05\\
623	5.52937998463676e-05\\
624	0.000112137470864582\\
625	6.11170881938961e-05\\
626	2.97983035579819e-05\\
627	2.5884735237014e-05\\
628	1.85937217338505e-05\\
629	2.48035986487594e-05\\
630	0.000326906627606149\\
631	1.42014046105032e-05\\
632	0.000182216818075176\\
633	1.83438746717002e-05\\
634	2.58459268707156e-05\\
635	2.42935566348024e-05\\
637	4.00238691327104e-05\\
638	7.08579534602904e-05\\
639	6.21494511960918e-05\\
640	8.42634697164613e-05\\
641	9.88395551075882e-05\\
642	0.000107415862999005\\
643	0.000125339382407219\\
644	0.00015556517200821\\
645	0.000199596912580091\\
646	0.00019218936336074\\
647	0.000231077127637036\\
648	0.000524074657391793\\
649	0.000272368794071554\\
650	0.000299483183084703\\
651	0.000492216530005862\\
652	0.000353984477328595\\
653	0.000438975487554671\\
654	0.000401138328356865\\
657	0.000490433159143322\\
658	0.000672532168891848\\
659	0.000549938746530643\\
660	0.000578509921638215\\
661	0.00061209229738779\\
662	0.000661269588704272\\
663	0.000677025263371434\\
664	0.000688616952629673\\
665	0.000713897154014376\\
666	0.000819581747350639\\
667	0.000802904613124156\\
668	0.000851148009019198\\
669	0.000803368550464256\\
670	0.000831007632655201\\
671	0.000851415650004128\\
672	0.00125114194440812\\
673	0.00086907288925886\\
674	0.000880942063645307\\
675	0.000896198544751314\\
676	0.000965735996715011\\
677	0.000993446398183551\\
678	0.000935079144596784\\
679	0.000915967551804897\\
680	0.000917387494268951\\
681	0.000932303825834424\\
682	0.000919507821663501\\
683	0.000920759023074424\\
684	0.000901159464433476\\
685	0.000901458135331268\\
686	0.000224258854359505\\
687	1.17840374969091e-06\\
688	0\\
689	1.7013312231029e-05\\
690	5.79852410282911e-05\\
691	7.65471829708986e-05\\
693	8.13830960169071e-05\\
694	0.000130040918065788\\
696	0.00014375162027136\\
697	0.000249795087711391\\
699	0.00025034363955001\\
700	0.000741413808782454\\
701	0.000771248408909742\\
702	0.000829680712058689\\
703	0.000950081849850903\\
706	0.00104384245662459\\
709	0.00114397867037042\\
712	0.00125793234713001\\
716	0.00137487682923165\\
717	0.00140034394746635\\
719	0.00146758487687748\\
720	0.00151277813165507\\
723	0.00159670385312438\\
724	0.00236497448463524\\
726	0.00233836877600879\\
728	0.00228950729945154\\
729	0.00226306053809822\\
730	0.00228577440236677\\
731	0.00226777181956095\\
732	0.00223804319762166\\
733	0.00223514594301832\\
735	0.00216194898382582\\
736	0.00211923575432138\\
737	0.00220605536048879\\
739	0.00211580404470624\\
740	0.00206988128013563\\
741	0.0021362908954643\\
742	0.00239783557731243\\
743	0.00239878101831437\\
744	0.00234164283085647\\
745	0.00229155387455648\\
746	0.0022352878052061\\
747	0.00219862209905841\\
748	0.00214904743302213\\
749	0.00210432612243494\\
750	0.00206651961991611\\
751	0.00203316883073512\\
752	0.00200572272444788\\
753	0.00198528151031496\\
754	0.0019712116370556\\
755	0.00222198984863553\\
756	0.00222056662357772\\
757	0.00222731084386396\\
758	0.00223843083756492\\
759	0.00227422486932483\\
760	0.00229251824828225\\
761	0.00232440838055936\\
762	0.00236469179287724\\
763	0.00240924203103532\\
765	0.00251973454282961\\
766	0.0025841580800261\\
767	0.00265713060798589\\
768	0.00273442933519163\\
769	0.00281704712847386\\
770	0.00290594945113298\\
771	0.00301558709429628\\
772	0.00309599734839594\\
773	0.00336221242821466\\
774	0.00330333648150827\\
775	0.00340967233273659\\
776	0.00351909016239915\\
777	0.00363333782172504\\
778	0.00377632030006225\\
779	0.00386369468526482\\
781	0.00408840377792785\\
782	0.00419859847704629\\
783	0.00450052519443034\\
784	0.00618160627482212\\
785	0.00452903582686304\\
786	0.00464579375739049\\
787	0.004735836128134\\
788	0.00489912913360513\\
789	0.00492237884748192\\
790	0.00500388412183383\\
791	0.00508154348767675\\
792	0.00516445890343675\\
793	0.00522686157182761\\
794	0.0052808472171364\\
795	0.00530991249581803\\
796	0.0053513743100666\\
797	0.00538105249631826\\
798	0.00516856458807524\\
799	0.00518883338861542\\
800	0.00520505433098606\\
801	0.00282429914602744\\
802	0.00284420033119659\\
803	0.00276045313319173\\
804	0.00295749665758649\\
805	0.00271453816810663\\
807	0.00266095229687835\\
808	0.00263215755001056\\
809	0.00260769697604246\\
810	0.00259451244255615\\
811	0.00255203300480389\\
812	0.00252803170008065\\
813	0.00247153847897152\\
814	0.00246307195959616\\
815	0.00238898231282292\\
816	0.00236495975650541\\
817	0.00236692931946436\\
818	0.00230130367560832\\
819	0.00224897673354008\\
820	0.00221815493159738\\
821	0.00221801764200791\\
822	0.00217823277375828\\
824	0.00214855132185221\\
825	0.00167620816091585\\
826	0.00168183863706872\\
827	0.00163822560386961\\
828	0.00140073882971592\\
831	0.00142694288524581\\
832	0.0013830523124625\\
834	0.00126927801716192\\
836	0.00127384362065186\\
837	0.00128045411599942\\
838	0.00113744850943931\\
839	0.00113926136089049\\
840	0.00115942714444373\\
841	0.00113942415202928\\
842	0.000875521021612258\\
843	0.000782456960337186\\
844	0.00074172665711103\\
845	0.000866189860971645\\
846	0.000686795724277545\\
847	0.000650568433229637\\
848	0.000648873280965745\\
849	0.000691150909233329\\
851	0.000584191396569622\\
852	0.000601409829187105\\
853	0.000626283444603359\\
854	0.000547681832244962\\
855	0.000535501340664268\\
856	0.000625404622837777\\
857	0.000452304657983404\\
858	2.34492517847684e-06\\
860	4.76214370337402e-06\\
861	1.29226867784382e-05\\
862	1.40582268386424e-05\\
863	6.37849251461375e-05\\
864	6.21058344449921e-05\\
865	8.26399573270464e-05\\
866	0.000133474147446577\\
868	0.000128929290667656\\
869	0.000135568756036264\\
870	0.000156427465981324\\
871	0.000153615258568607\\
872	0.000223061242081712\\
873	0.000284886741269474\\
874	0.000281080055174243\\
875	0.000323255963280644\\
877	0.000310096391444858\\
879	0.00030143432445584\\
880	0.000299126796562632\\
881	0.000348986308381427\\
882	0.000379400496058224\\
884	0.000383154502287653\\
886	0.000393927970662844\\
887	0.000411376455531354\\
888	0.000620545600668265\\
889	0.000708456102870514\\
890	0.000760228371746052\\
891	0.000773904319544272\\
892	0.000794586581150725\\
893	0.000810416597005315\\
894	0.000837416841591221\\
895	0.00085948736489172\\
896	0.00151402090659758\\
897	0.0014940762275728\\
899	0.00146057117399323\\
901	0.00144439762982529\\
903	0.00145302584644469\\
904	0.00151640498961569\\
905	0.00154854912159408\\
906	0.00156290868039832\\
907	0.00158255830183407\\
908	0.00165071639833059\\
909	0.00168222988668276\\
910	0.00171827424219373\\
911	0.00175814623662518\\
912	0.00180229034276636\\
913	0.00184990590662437\\
914	0.00191045055987615\\
916	0.00202127677903263\\
917	0.00434515710264805\\
919	0.00462784063608979\\
921	0.00491816375108556\\
922	0.0050640632739487\\
924	0.00533123778245681\\
925	0.00548341914111461\\
928	0.00582638173568739\\
929	0.00592701544076135\\
930	0.00604790025090551\\
931	0.00615160611437204\\
932	0.00631827973313648\\
933	0.00636492106934838\\
935	0.00643790592073401\\
936	0.00653272080842271\\
937	0.00653244206523595\\
939	0.00650892154772009\\
940	0.00655242994344007\\
941	0.00650159397980588\\
942	0.00644338062306815\\
943	0.00637885908690805\\
944	0.00630788990974906\\
946	0.00613108881918834\\
947	0.00606486658375616\\
948	0.00595771612222507\\
949	0.00583299591255582\\
950	0.00571251001679229\\
951	0.00557776162418122\\
954	0.00514865648381146\\
955	0.00499661377853045\\
956	0.00484826836282082\\
957	0.00470654505261336\\
958	0.00455379925597299\\
960	0.00427100244007761\\
962	0.00398803045925433\\
963	0.00385988238326718\\
964	0.00372547378583477\\
965	0.00364246343815466\\
966	0.00352738310300538\\
967	0.00310946821264224\\
968	0.00300392985025155\\
969	0.0029017360556054\\
970	0.00280926282778182\\
971	0.00272495625995361\\
973	0.00257581041284993\\
974	0.00251946289597527\\
975	0.00247493851463787\\
976	0.0024251849350776\\
977	0.0023958377270219\\
978	0.00237724706039444\\
979	0.00236645807592595\\
980	0.00241717781239004\\
981	0.00251808219502436\\
982	0.00252064095241167\\
984	0.00254896216131328\\
985	0.00258212495441512\\
986	0.00261066855705394\\
987	0.00266967906156879\\
988	0.00269625891542091\\
989	0.00244315097074832\\
990	0.00248054043231605\\
992	0.00263819403176058\\
993	0.00271383711651652\\
994	0.00279476673210866\\
995	0.00284855273741869\\
996	0.00292611416944055\\
997	0.0025581514194073\\
998	0.00264105744395238\\
999	0.00271580093021839\\
1000	0.00278066967166524\\
};
\end{axis}

\begin{axis}[%
width=4.875in,
height=2.036in,
at={(0.758in,1.1in)},
xmin=0,
xmax=1000,
xlabel style={font=\bfseries\color{white!15!black}},
xlabel={Time in slots},
ymin=0,
ymax=0.04,
axis background/.style={fill=white},
title style={font=\bfseries},
title={(b) Interference with channel dynamics only},
axis x line*=bottom,
axis y line*=left,
xmajorgrids,
ymajorgrids
]
\addplot [color=blue, forget plot]
  table[row sep=crcr]{%
1	0.00992780891067468\\
3	0.00981372578416995\\
6	0.00965945416191971\\
10	0.00954504590617944\\
12	0.0094980206343962\\
13	0.00950145712488393\\
15	0.00946663219087895\\
23	0.00942240167319142\\
26	0.0093720396603203\\
28	0.00932725454981664\\
30	0.00926984820250709\\
32	0.0092129452153813\\
34	0.00914132212540153\\
35	0.00908301521110388\\
36	0.00900866027996017\\
37	0.00895149560949449\\
40	0.00872313638285505\\
42	0.00853928613594235\\
47	0.00804455585171127\\
49	0.00784368518168321\\
52	0.00753821267733201\\
54	0.00736712745333534\\
55	0.00727927178900245\\
57	0.00716308569440116\\
59	0.00707439643201724\\
61	0.00703990860483827\\
62	0.00702624019913856\\
64	0.00704362269925696\\
66	0.00713410306821061\\
68	0.00726557546920503\\
69	0.00735662824229166\\
70	0.00746428390345955\\
72	0.00772073685914165\\
74	0.00804607903717169\\
75	0.00822256909418684\\
77	0.00863064109739753\\
78	0.00885200785546658\\
80	0.00933493386150985\\
81	0.00959206832840209\\
82	0.00986918172793594\\
86	0.0109116436893828\\
88	0.0114960697394508\\
89	0.0117446356871369\\
94	0.0131507577799539\\
96	0.0135605975043518\\
97	0.0137418087847436\\
98	0.013959703479145\\
99	0.0141558504211616\\
100	0.0143059484435071\\
101	0.0144399379414608\\
102	0.0145251650262708\\
105	0.0147044449712439\\
106	0.0146969251819655\\
107	0.0147231291133494\\
108	0.0147076187490711\\
111	0.0145492584308613\\
112	0.014401813038603\\
113	0.0142216174451733\\
115	0.0139273984182182\\
118	0.0132855841366109\\
119	0.0130165133012952\\
120	0.0127805072776255\\
121	0.0124701787004824\\
122	0.0121751257139522\\
124	0.0116211390705985\\
125	0.0113260192582629\\
128	0.0103156963942865\\
129	0.0100180406413983\\
130	0.00967989580442463\\
134	0.00849904057679396\\
137	0.00770994532570057\\
139	0.0072754601521865\\
141	0.00687816400522934\\
142	0.00670168405099503\\
143	0.00655706406223544\\
145	0.00631500160386622\\
146	0.00622221070295836\\
147	0.00615119222709382\\
149	0.00606836497217955\\
150	0.00604467970856604\\
151	0.00604809084632052\\
152	0.00606774186042003\\
154	0.00615520553276383\\
155	0.00621965309346706\\
157	0.00639978845504174\\
160	0.00671665181278058\\
163	0.00709768348019679\\
167	0.00758378413502214\\
171	0.00797603086743948\\
173	0.00815364966842935\\
176	0.00826801841890301\\
177	0.00827026140211728\\
178	0.00829557118390767\\
180	0.00830388061081067\\
184	0.00807328533994678\\
186	0.00792452056498405\\
188	0.00771333689397125\\
189	0.00758041508186125\\
191	0.0073581216410048\\
192	0.00721735890908803\\
197	0.00663709691741587\\
200	0.00639047626043521\\
201	0.00632086346081451\\
203	0.00625016571530068\\
204	0.00624128365029719\\
206	0.00629972432386694\\
208	0.00645324490494659\\
209	0.00656429351397492\\
210	0.00670922914412131\\
211	0.00687826324565322\\
212	0.00707973442160892\\
213	0.00729862429716377\\
214	0.00754834848225983\\
215	0.00782277107191476\\
216	0.0081403400971567\\
217	0.00847411704921797\\
218	0.00884206866146542\\
219	0.00923141892985768\\
220	0.00965401168366498\\
222	0.0105618655829858\\
224	0.0115622640656738\\
228	0.0137766530136787\\
229	0.0143652122760614\\
230	0.0149203460059653\\
231	0.0155256342120538\\
232	0.0160988930798567\\
233	0.0166891507191167\\
234	0.0172489690836528\\
235	0.0178668880339501\\
236	0.0184095623591247\\
237	0.0190079492464292\\
238	0.0195731720131107\\
239	0.0201057882696887\\
240	0.0206199323150713\\
241	0.0211669234387273\\
243	0.0221804701606061\\
244	0.0225867112062588\\
245	0.0229711062549995\\
246	0.0233961812626831\\
247	0.0237111656671232\\
248	0.0240087725566127\\
249	0.0242793231587939\\
250	0.024527619883429\\
251	0.0247518123544523\\
252	0.0250316832813269\\
253	0.025193393804102\\
254	0.0253363585021589\\
256	0.0254442168244395\\
258	0.0254514016379517\\
260	0.0254700225596025\\
261	0.0253807073681855\\
263	0.0252485612248847\\
264	0.0250708631515408\\
265	0.0249319043963396\\
267	0.0245792075604641\\
268	0.0244602952401465\\
269	0.0242476513831207\\
270	0.0241145626931711\\
271	0.0239100145877273\\
272	0.0237322133599491\\
273	0.0235851597205965\\
274	0.0233488615484703\\
275	0.0231994742666757\\
276	0.0230679686129633\\
277	0.0229108661692408\\
279	0.0227094514680175\\
281	0.0224359527912839\\
282	0.0223213909886226\\
283	0.0221901015886488\\
284	0.0221101182748953\\
285	0.0220128988013357\\
286	0.0220149523059945\\
288	0.0221096252520283\\
289	0.0222024701491819\\
290	0.0222783913208104\\
291	0.0223844316078612\\
292	0.0224582682441223\\
294	0.0226495688540354\\
295	0.0228188859521197\\
296	0.0229387052853554\\
297	0.0231301414917198\\
298	0.0232870514811339\\
299	0.023549112898877\\
300	0.0237712229428553\\
301	0.023954258800245\\
302	0.0242321327338004\\
303	0.0244219341478811\\
304	0.0246639037590057\\
305	0.0248152322322994\\
306	0.025102929146442\\
307	0.0252899602656953\\
308	0.0255512011295878\\
309	0.0257818133469527\\
310	0.0259381526038851\\
311	0.026137869807485\\
312	0.0262854536795203\\
313	0.0265107585088344\\
314	0.026714918636344\\
315	0.0267816578520979\\
316	0.0269571174060275\\
317	0.0269722904007494\\
318	0.0270336102139481\\
319	0.0270064001466608\\
321	0.0269917978756666\\
322	0.0270181395674172\\
323	0.0269376000651391\\
324	0.026892540984818\\
325	0.0266798135132831\\
327	0.0263901284641861\\
328	0.0262287830767036\\
329	0.0258930595191487\\
330	0.025584984174543\\
331	0.0253371743257276\\
332	0.0249541866409118\\
333	0.0246668550959157\\
334	0.0243451840268563\\
335	0.0238963016101934\\
336	0.0234729068415618\\
337	0.0230785052922329\\
338	0.0225913191898144\\
340	0.0216808849932022\\
341	0.0211520179840363\\
342	0.0206594213590279\\
343	0.0201901422384481\\
346	0.0186867669788171\\
347	0.0181741944779787\\
348	0.0176853970544926\\
351	0.0161743071415685\\
353	0.0152558384754684\\
354	0.0148633077034219\\
355	0.0144453167462189\\
356	0.0140512849573042\\
358	0.0133144583111289\\
359	0.0129812820745201\\
360	0.0126681973176801\\
362	0.0121871091903358\\
363	0.0119944290423746\\
364	0.0118408631025204\\
366	0.0115868781391555\\
367	0.0115108812287872\\
369	0.0113957658022628\\
370	0.0113658389806233\\
372	0.0113830099521692\\
375	0.011613024444955\\
377	0.0118236572740216\\
378	0.0119689745999949\\
379	0.0121341708045293\\
380	0.0122840801896018\\
381	0.0124714570074502\\
382	0.0126316533406907\\
384	0.0130196212899136\\
385	0.0132309064682659\\
386	0.0134071435890064\\
387	0.013615083100035\\
388	0.0137731981957359\\
390	0.0141440883161295\\
394	0.014768317671269\\
396	0.0149979486313896\\
398	0.0151984988715412\\
399	0.0153089474684975\\
402	0.0154867837673009\\
403	0.0155110681689621\\
405	0.0154699774626579\\
407	0.0154178793960682\\
409	0.0153486781107404\\
412	0.0152360507825051\\
413	0.0151798783530239\\
415	0.0149949204283075\\
416	0.0149226747201965\\
418	0.0147137997774962\\
420	0.0145556646408522\\
421	0.0144647757258554\\
422	0.0144005746886933\\
424	0.0142289543763354\\
425	0.0141684376827698\\
427	0.0140883027613654\\
430	0.0140392827429423\\
431	0.0140456216597613\\
432	0.0140749706629322\\
433	0.0140880883124055\\
435	0.0141608165831713\\
437	0.0142244655427248\\
438	0.0143131321207193\\
439	0.0143796049669618\\
440	0.0144996047951054\\
441	0.0145681278729626\\
442	0.0146603912294268\\
444	0.0149300723817305\\
445	0.0150563903275724\\
446	0.015164716598747\\
453	0.0157626594024123\\
454	0.0158568769106751\\
455	0.0158959915247578\\
456	0.0159162260588346\\
458	0.0160630653542739\\
459	0.0160763511738651\\
461	0.0161315841173746\\
463	0.0160003332941869\\
465	0.015750099948491\\
466	0.0156719707943012\\
468	0.0153669112434045\\
469	0.0152316822801595\\
471	0.0148050688310377\\
472	0.014610440068509\\
473	0.0143196057709929\\
474	0.0140556294480803\\
476	0.0134039472376344\\
477	0.0131317430993931\\
479	0.012471675883944\\
480	0.0121153313648392\\
481	0.011808627293135\\
482	0.0114686510270303\\
484	0.0108394986045823\\
485	0.0104884603565552\\
488	0.00952626377818433\\
491	0.00868727333261177\\
494	0.00795506257782108\\
495	0.00775897522692048\\
496	0.00759045436518591\\
497	0.00745110120772097\\
498	0.0073278115788753\\
499	0.00722771410210044\\
500	0.00715925216627511\\
501	0.00713687619611392\\
502	0.00714217829386143\\
504	0.00724349634094779\\
505	0.00734783154632623\\
506	0.00747917680337196\\
507	0.00764005608505158\\
508	0.00783285571026227\\
509	0.00806368665223545\\
510	0.00833917165346065\\
512	0.00893207620651992\\
514	0.00963780816755389\\
516	0.0104111828491114\\
517	0.0108414583465901\\
521	0.0126974575185841\\
526	0.0152310898158703\\
527	0.0157047168987674\\
529	0.0167028580609667\\
531	0.0177071736983407\\
532	0.0181631017073869\\
533	0.0185593562652002\\
535	0.0194241319574076\\
536	0.0197368582679474\\
537	0.0200221796916367\\
538	0.0202903208080443\\
539	0.0205069743457216\\
540	0.0207905280914247\\
541	0.0210153543323486\\
542	0.021256873122411\\
543	0.0213806544203408\\
544	0.0215689836909405\\
545	0.0216417094868575\\
546	0.021760539374668\\
548	0.0218856095439151\\
549	0.0218853210640191\\
551	0.0218330330211529\\
552	0.0217661959422912\\
553	0.0216492980757721\\
554	0.0215162244136309\\
555	0.0214338816007285\\
556	0.0213003468047646\\
559	0.0207361179407144\\
562	0.0201183312688045\\
566	0.0193206418284717\\
568	0.0189494335185145\\
569	0.0187403719384065\\
570	0.0185472440424519\\
571	0.0183921757309236\\
572	0.0182613956557134\\
573	0.0181012810818402\\
575	0.0178398807381654\\
576	0.0177153109673327\\
578	0.0175253284160135\\
579	0.017442648935571\\
580	0.0173963405901532\\
581	0.0174149170248938\\
582	0.0173892593629716\\
583	0.0173828173984703\\
585	0.01744044840882\\
586	0.0175143434188385\\
588	0.0175956146327962\\
589	0.0176520034008263\\
590	0.0177284745935822\\
591	0.0178290206252996\\
592	0.0179049724920333\\
594	0.0181460040026877\\
595	0.0182515565410313\\
596	0.0183215156548613\\
598	0.0185304354585014\\
599	0.0185915804563592\\
600	0.018677584166312\\
603	0.0187138781407157\\
604	0.0186857172503778\\
606	0.0186025714824609\\
607	0.018577149264388\\
609	0.0184451318040146\\
610	0.0182954472437586\\
611	0.0181841326327685\\
613	0.0178870545238397\\
614	0.0176663374895725\\
616	0.0172441487154629\\
617	0.0169760630443534\\
618	0.0167387393164518\\
619	0.0164601499998298\\
621	0.0158032850953305\\
622	0.0154674403664785\\
623	0.0151840086563197\\
628	0.0136213703067369\\
631	0.0127450075169691\\
633	0.0122281161758337\\
634	0.0119634725685955\\
635	0.0117296381184815\\
636	0.0115206109459223\\
637	0.0113349192828309\\
638	0.0111682994698867\\
640	0.0109175019289296\\
641	0.0108459387040512\\
642	0.0108045074462098\\
644	0.0108093596020353\\
645	0.0108801910733973\\
647	0.0110961805788747\\
648	0.011266174651837\\
649	0.0114726081696972\\
650	0.0117028939075681\\
651	0.0119572366880902\\
652	0.0122601293459184\\
654	0.0129743718285908\\
657	0.0141394039923171\\
658	0.0145672460338346\\
659	0.0150280650752848\\
660	0.0154716691092744\\
661	0.0159905553005046\\
662	0.0165574443758487\\
663	0.0170499113503411\\
664	0.0175884922789464\\
665	0.0180902096035425\\
666	0.0186536405133211\\
667	0.0191291298827991\\
669	0.0202799405828955\\
671	0.0214116669453688\\
672	0.0218850308984884\\
673	0.0222874284637555\\
674	0.0227341685648526\\
676	0.0234281313074689\\
677	0.0238038773380822\\
678	0.0241235116101279\\
679	0.0243805893971967\\
680	0.0246791810363902\\
681	0.0249623076526859\\
682	0.0251998116309551\\
683	0.0253071931281283\\
684	0.0253760882901588\\
685	0.0255134657554663\\
686	0.0256100133759674\\
687	0.0255782400377029\\
689	0.0253770708106913\\
690	0.0252118604638554\\
691	0.025069246401813\\
692	0.0248833232559491\\
693	0.0247155475976797\\
694	0.0244024299103103\\
695	0.0241162186445081\\
697	0.0234433407780443\\
698	0.0230101975977277\\
699	0.0226315260651972\\
700	0.0221295266852621\\
701	0.0216605164694101\\
702	0.0211290214103883\\
703	0.0205631823505428\\
704	0.0200821123718242\\
705	0.0195717533267725\\
706	0.019025836404353\\
708	0.0180035836843899\\
709	0.0174449947026005\\
711	0.0164399840136866\\
713	0.0153847052058609\\
714	0.0149129792998792\\
715	0.0144754204370656\\
716	0.0140587473347296\\
717	0.0136116088400513\\
720	0.0124102901695551\\
721	0.0120592274812452\\
723	0.0114562436732513\\
724	0.0111998348185125\\
725	0.0110108976839456\\
726	0.0108028879875519\\
727	0.0106411872847048\\
728	0.0105336667270421\\
729	0.0104573341849346\\
730	0.0104068639062689\\
731	0.0103745015322829\\
732	0.0103908608326719\\
733	0.0104436395744187\\
734	0.0104774216456462\\
736	0.010680749032872\\
737	0.0108384903456908\\
738	0.0110237524196464\\
739	0.0111859438081865\\
741	0.0115646610573776\\
742	0.0117768524418125\\
743	0.0120113127553623\\
744	0.0122850178657927\\
745	0.0125352834679688\\
747	0.0131001729145055\\
750	0.0140705325524095\\
751	0.0143708575239998\\
752	0.0147196679242825\\
753	0.0150520280564024\\
754	0.0153610194080329\\
755	0.0156294319996277\\
756	0.0159710156375468\\
758	0.0166112840751111\\
759	0.0168442257825063\\
762	0.0177084009636701\\
765	0.0182718265851918\\
767	0.0186886792129144\\
768	0.018819922883722\\
769	0.0189732816147625\\
770	0.0191855967962056\\
771	0.0193681398903891\\
772	0.0194561519829222\\
773	0.0196170848887505\\
774	0.0196997167053041\\
775	0.0198289392443485\\
777	0.0200082853822323\\
778	0.0200523657518943\\
780	0.0202167577012915\\
781	0.0202913936190043\\
782	0.0203984543562683\\
783	0.0203848054302398\\
785	0.0205118508822579\\
787	0.0205129286356396\\
789	0.0204751655677455\\
790	0.0204448972439195\\
791	0.0204508692098671\\
792	0.0205058408467949\\
793	0.0204724731764827\\
794	0.0204995731347708\\
795	0.0204855520154297\\
796	0.0205206120490402\\
797	0.0205034823804908\\
798	0.0204682477490223\\
799	0.020526144660721\\
800	0.0205471370840087\\
802	0.0205423511690697\\
803	0.0205576769266145\\
804	0.0205162248105353\\
806	0.0204979400544971\\
807	0.0204616399214501\\
809	0.0205618367763236\\
810	0.0205360673822952\\
811	0.0205390383694066\\
812	0.0205737741019902\\
813	0.0205164497691612\\
815	0.0205437593434681\\
816	0.0204943977076937\\
817	0.0204879083555625\\
818	0.0204301533268563\\
819	0.0204411094304078\\
820	0.0203636916098731\\
821	0.0203276603737095\\
822	0.0202756002379374\\
823	0.0202080733008643\\
825	0.0201019721677085\\
826	0.0199795104998657\\
827	0.0198938818331271\\
828	0.0197762715571344\\
829	0.019685116343112\\
831	0.0194435290421779\\
832	0.0192919486457868\\
834	0.0190659462386975\\
835	0.0189283201430044\\
836	0.0187127506778779\\
838	0.0183701685042479\\
839	0.0181842682494562\\
841	0.0177697433824733\\
842	0.0174996347045635\\
844	0.0169971875277497\\
845	0.016717590873327\\
846	0.0165009228605868\\
847	0.0162665263251256\\
849	0.0157470806191213\\
850	0.0154438980465557\\
851	0.0152028064252363\\
852	0.0148979406474155\\
853	0.0146193327835817\\
854	0.0143216428839423\\
855	0.0140437659302961\\
857	0.0134584261435293\\
858	0.0131902142090894\\
861	0.0124684087011246\\
862	0.0122678047608815\\
863	0.012028693252546\\
866	0.0115228706935113\\
869	0.0111121968908492\\
871	0.0109241321861191\\
872	0.0108987374474054\\
873	0.010850385208073\\
874	0.0108593574124143\\
875	0.0109008071420931\\
876	0.0109738267981356\\
877	0.0110668490666512\\
879	0.0113477968207008\\
881	0.0117052806421043\\
882	0.0119159914393094\\
883	0.0121911203166292\\
885	0.0127875618411508\\
888	0.0139625869039719\\
889	0.0144038112617864\\
890	0.014863575606114\\
891	0.0153399736417441\\
892	0.0158884904236629\\
893	0.0164525241692672\\
896	0.0182233533895442\\
897	0.0188791145494633\\
898	0.019513134114618\\
900	0.020899459462612\\
901	0.021642295640504\\
902	0.0223690833515775\\
903	0.0231316078180726\\
904	0.0238549699218993\\
905	0.0246390669184393\\
906	0.025387356100282\\
907	0.0261579540341472\\
908	0.0269562942676203\\
909	0.0276710646122638\\
910	0.0284716255953299\\
911	0.0293295332180605\\
912	0.0300578386517145\\
913	0.0308039603339694\\
914	0.031604894584575\\
915	0.0322639968102294\\
916	0.032945257926599\\
917	0.0335643805263999\\
918	0.034267321044922\\
919	0.0348938297171344\\
920	0.0353610367810688\\
921	0.0358774274614007\\
922	0.0362597723200224\\
923	0.0365914051278651\\
924	0.0368991651727129\\
925	0.0372507710707168\\
926	0.0376193900593762\\
927	0.0377968402569877\\
928	0.0381165096788436\\
929	0.0383453007912067\\
930	0.038541956188169\\
931	0.0386307398960071\\
932	0.0387383754691655\\
933	0.0386482061570632\\
934	0.0384967549662178\\
936	0.0381309217706303\\
937	0.0377629054872841\\
938	0.0375152076589984\\
939	0.0372149535196513\\
941	0.0364944760839307\\
943	0.0356124197155623\\
944	0.0349854180790317\\
945	0.0342955265450655\\
946	0.0335429944930183\\
948	0.0323253019705589\\
949	0.0315027687150859\\
950	0.0308455561183791\\
951	0.0301619703296865\\
952	0.0293462694146456\\
954	0.0276424088350495\\
955	0.0268102908078163\\
956	0.0260153777915093\\
957	0.0252689125543384\\
958	0.0244320285825097\\
959	0.0236188339639511\\
960	0.0227861068609627\\
961	0.02208512330742\\
963	0.0207164635243089\\
964	0.0199935443534969\\
965	0.019299158555782\\
966	0.0186395042463801\\
967	0.018089463201818\\
969	0.0170365728790784\\
970	0.0165557093361031\\
971	0.016111319569859\\
972	0.0157552266631455\\
973	0.0154263607009852\\
974	0.0151619056663321\\
975	0.0149273831387973\\
976	0.0147218139986762\\
977	0.0145670449161344\\
978	0.0144873742857499\\
979	0.0144422974281042\\
981	0.0144917972558005\\
982	0.014584766449957\\
983	0.0146997508558115\\
984	0.0148504579126438\\
985	0.0150342568435917\\
987	0.0155591176638836\\
988	0.0158861924182929\\
989	0.0162286813210812\\
990	0.0165890227483487\\
993	0.0178512252446126\\
994	0.0183223130659371\\
998	0.0200539865270457\\
1000	0.0209839938697769\\
};
\end{axis}

\begin{axis}[%
width=4.875in,
height=3.375in,
at={(0.6in,1.1in)},
scale only axis,
xmin=0,
xmax=1,
ymin=0,
ymax=1,
ylabel style={font=\bfseries\color{white!15!black}},
ylabel={Interference power in mW},
axis line style={draw=none},
ticks=none,
axis x line*=bottom,
axis y line*=left
]
\node[below right, align=left]
at (rel axis cs:0.05,1.02) {$\mu= 0.01$, $\ell=10$, $\eta=50$};
\node[below right, align=left]
at (rel axis cs:0.05,0.38) {$\mu= 1$, $\eta=50$};
\end{axis}
\end{tikzpicture}%
}   
\caption{Interference traces depicting the impact of traffic on the interference dynamics. Part (a) shows a realization of Setup~1 from Table~\ref{tab:Scenarios}. Part (b) shows a realization of Setup~1 that ignores the traffic contribution ($\txp=1$), only accounts for the channel dynamics. Transmit power $\txpwr=1$~mW.} 
\label{Fig:Interference_traces}
\end{figure}

Fig.~\ref{Fig:Interference_traces} illustrates the impact of different correlation sources on the overall interference. The upper plot shows the interference resulting from time-varying traffic over a time-varying channel. The lower plot shows the interference if all nodes transmit all the time over the same time-varying channel. The  differences are visible and underline the limitations of channel prediction in an interference-dominated system. 

We design an interference predictor based on the Kalman filter that harnesses the knowledge of the correlation  \eqref{eq:AutocorrFunction} and is practically feasible in a sense that nodes can implement it with limited computational resources and without additional signaling. 
The interference in \eqref{eq:interf_def} is a nonlinear function of the considered sources of temporal correlation. Nonlinear approaches, like the extended or unscented Kalman filter, would be a natural choice for deriving an interference predictor. However, aiming at a predictor for resource constrained devices, we leave the use of nonlinear filters for future~work. 

\begin{figure*}[!t]\centering
	\includegraphics[width=\textwidth,keepaspectratio=true]{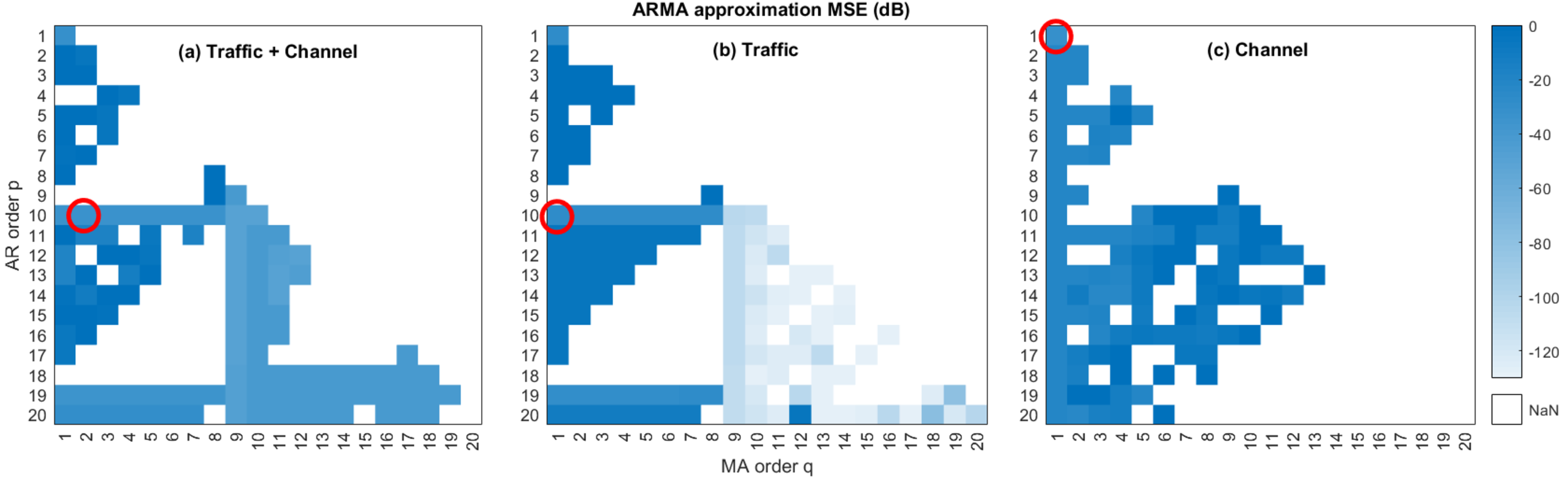}
\caption{Heatmaps showing the ARMA approximation MSE (dB) achieved by different $(p,q)$-pairs on three autocorrelation functions. Part (a) corresponds to Setup~1 from Table~\ref{tab:Scenarios}, part (b) to Setup~1 simplified to have time-invariant channel conditions, and part (c) to the simplification of Setup~1 to have time invariant traffic. The selected model orders, meeting an MSE target of $-30$~dB, are highlighted with red circles.} 
\label{Fig:ApproxError}
\end{figure*}  

\subsection{ARMA Approximation of $\rho(\tau)$}
\label{subsec:ARMAapproc}

ARMA models are extensively used in applications involving temporal stochastic processes. The Cram\'er-Wold theorem \cite[Ch.~17]{Pollock} states that every stationary stochastic process has a moving average (MA) representation. If an auto\-regressive (AR) component is also used, the potential to accurately represent the process with a limited number of parameters greatly increases, making ARMA models particularly useful.

The ARMA representation of a process is typically identified from observing its realizations; the autocorrelation of the process is in general unknown. Such an identification approach requires several observations to identify the model dimension and its coefficients'\:values. These observations constitute a form of pilot signaling and can become impractically large in some cases, depending on the dynamics of the process to represent. 
Different to the typical case, we know $\rho(\tau)$ for a given system model in terms of the sources of correlation considered in \eqref{eq:AutocorrFunction}. We are therefore interested in approximating it through an ARMA model with a small number of parameters. 

The ARMA($p$, $q$) representation of $i(t)$ can be written as
\begin{equation}
	\sum_{n=0}^p a_n \, i(t-n) = \sum_{n=0}^q b_n\,\epsilon(t-n), \qquad q\leq p,
	\label{eq;ARMAdef}
\end{equation}
where the model orders $p$ and $q$ are in general unknown. The coefficients $a_n$ and $b_n$ specify the AR and MA components, respectively, and $\epsilon(t)$ is a zero-mean white noise. The model is normalized by setting $a_0 = b_0 = 1$. 

The coefficients in \eqref{eq;ARMAdef} can be inferred from $\rho(\tau)$. Specifically, multiplying \eqref{eq;ARMAdef} by $i(t-\tau)$ and taking expectation~yields
\begin{equation}
	\sum_{n=0}^p a_n\,\rho(\tau-n) = \sum_{n=0}^q \,b_n\delta(n-\tau),
	\label{eq;ARMAdef2_aux}
\end{equation}
with the Dirac delta function $\delta(\cdot)$. 
Specializing for $\tau=q+1, \ldots, q+p$, the AR coefficients $a_1,\ldots,a_p$ can be found as the solution to the resulting Yule-Walker equations \cite{RegaliaLibro} (see Appendix~A). 

To determine the MA coefficients, an auxiliary sequence $\psi(t)$ defined to equal each side of \eqref{eq;ARMAdef} is introduced. Then: 
\begin{eqnarray}
	\label{eq:bn_extraction1_aux}
	\expect{\psi(t)\,\psi(t-\tau)}\!\!&=&\!\!\sum_{m=0}^p\sum_{n=0}^p a_m\,a_n\,\rho(\tau\!+\!n\!-\!m) \\ 
		&=&\!\!\sigma_\epsilon^2\sum_{m=0}^q b_n\,b_{n+\tau}.
	\label{eq:bn_extraction2_aux}
\end{eqnarray}
Since all terms on the right hand side of \eqref{eq:bn_extraction1_aux} are known, the terms $b_1, \ldots, b_q$ and $\sigma_\epsilon^2$ can be solved for (see Appendix~A) by equating that expression to \eqref{eq:bn_extraction2_aux}.

Observe that in \eqref{eq;ARMAdef2_aux} and \eqref{eq:bn_extraction1_aux} only the first $p+q$ values of $\rho(\tau)$ are involved in the computation of the model coefficients. For a good representation, $p+q$ should be large enough to capture the main features of $\rho(\tau)$. If significant correlation exists for large $\tau$-values, a decimation of $\rho(\tau)$ can be performed to reduce the number of significant lags prior to the parameter determination to keep $p+q$ low. A scaling down in frequency of the resulting ARMA model in the same decimation factor is then performed to bring the~result to the same scale. 

\subsection{ARMA Model Order Selection}
\label{subsec:ARMAorder}

The described coefficient determination depends on the orders $p$ and $q$ and is thus not unique. Each $(p, q)$-pair results in an approximation error for $\rho(\tau)$; many pairs yield low errors, but not all pairs are suitable. 
The availability of $\rho(\tau)$ allows a least square approach to select the order values. Feasible $(p, q)$-pairs are those for which $\hat{\rho}_{p,q}(\tau)\rightarrow\rho(\tau)$ for $\tau\rightarrow\infty$. For this set and a largest significant correlation lag of $T$, the mean square approximation error is
\begin{equation}
	\label{eq:ApproxError}
	{\rm MSE}(p,q) = \frac{1}{T} \sum_{\tau=1}^T \Big(\rho(\tau)-\hat{\rho}_{p,q}(\tau)\Big)^2.
\end{equation}
Pair $(p,q)$ is then chosen as that lowest order-$p$ model meeting a target error. This selection criteria is illustrated in Fig.~\ref{Fig:ApproxError} for three different autocorrelation functions derived from Table~\ref{tab:Scenarios}, considering $T\!=\!100$ and a model order of up to $p=20$. These examples show how the order is affected by the different sources of correlation.

To obtain the errors in Fig.~\ref{Fig:ApproxError}, it is necessary to find the autocorrelation $\hat{\rho}_{p,q}(\tau)$ in \eqref{eq:ApproxError} associated to a given $(p,q)$-pair, which we simply call $\hat{\rho}(\tau)$. Using $\hat{\rho}(\tau)$ in \eqref{eq;ARMAdef2_aux} and noting that $\delta(n-\tau)=\expect{i(t-\tau)\epsilon(t-n)}=0$ for $\tau>n$, it follows that
\begin{equation}
	\label{eq:Coeff2Corr}
	\sum_n a_n \, \hat{\rho}(\tau-n) = 0 \quad \mbox{for} \quad \tau>q.
\end{equation}
Thus, knowing the $q+1$ nonzero values of $\delta(\cdot)$ and the $p$ initial values of $\hat{\rho}(\cdot)$, \eqref{eq:Coeff2Corr} can be solved recursively for all values of $\hat{\rho}(\cdot)$ beyond $p$ that are needed to compute~\eqref{eq:ApproxError}. 

To obtain the required nonzero $\delta$-values, we follow \cite[Ch.~17]{Pollock}. Multiplying \eqref{eq;ARMAdef} by $\epsilon(t-\tau)$ and taking expectation yields
$\sum_{n=0}^\tau a_n \, \delta(\tau-n) = \sigma_\epsilon^2 \, b_\tau$, which can be rearranged,
\begin{equation}
	\label{eq:Coeff2Corr_3}
	\delta_\tau = \frac{1}{a_0}\left(b_\tau\sigma_\epsilon^2-\sum_{n=0}^\tau a_n \delta(\tau-n)\right),
\end{equation}
to recursively find $\delta(0),\delta(1),\ldots,\delta(q)$. The $p$ initial values of $\hat{\rho}(\cdot)$ are obtained by substituting $\delta(0),\delta(1),\ldots,\delta(q)$ and $\hat{\rho}(\cdot)$ in \eqref{eq;ARMAdef2_aux}. Finally, the initial values of $\hat{\rho}(\cdot)$ are used to compute the succeeding ones using \eqref{eq:Coeff2Corr}.

\subsection{Recursive Interference Predictor}

\begin{figure*}[!t]
 \centering
  \resizebox{\textwidth}{!}{
  \definecolor{mycolor1}{rgb}{0.39216,0.83137,0.07451}%
\begin{tikzpicture}

\begin{axis}[%
width=3.382in,
height=3.5in,
at={(5.5in,0.573in)},
xmin=1,
xmax=10,
xlabel style={font=\bfseries\color{white!25!black}},
xlabel={Prediction horizon in slots},
ymin=-8,
ymax=3,
axis background/.style={fill=white},
title style={font=\bfseries},
title={(b) Setup 2},
axis x line*=bottom,
axis y line*=left,
xmajorgrids,
ymajorgrids,
legend style={at={(0.45,0.05)}, anchor=south west, legend cell align=left, align=left, draw=white!15!black}
]
\node[below right, align=left]
at (rel axis cs:.05,0.96) { $\eta = 20$ slots\\($\nu = 0.0191$ m/slot)};
\addplot [color=blue, line width=1.0pt, mark size=4.0pt, mark=o, mark options={solid, blue}]
  table[row sep=crcr]{%
1	-7.24800712875295\\
2	-4.93320924308036\\
3	-3.30054124856727\\
4	-2.51394105659232\\
5	-1.75102890720539\\
6	-1.35898998572295\\
7	-1.04250956584114\\
8	-0.806380270384727\\
9	-0.672317447959351\\
10	-0.63926053583379\\
};
\addlegendentry{Interference predictor}

\addplot [color=red, line width=1.0pt, mark size=4.0pt, mark=x, mark options={solid, red}]
  table[row sep=crcr]{%
1	-7.37946181507515\\
2	-4.5830064670155\\
3	-2.83860605472544\\
4	-1.72985929785883\\
5	-0.812974617938083\\
6	-0.0634678701898306\\
7	0.582250632524799\\
8	1.29184994737523\\
9	1.70288019893825\\
10	1.98627652479895\\
};
\addlegendentry{Channel predictor}

\addplot [color=mycolor1, line width=1.0pt, mark size=4.0pt, mark=+, mark options={solid, mycolor1}]
  table[row sep=crcr]{%
1	-7.51881162926411\\
2	-4.55685857751717\\
3	-2.79192522861081\\
4	-1.57510783088961\\
5	-0.617390908247298\\
6	0.136657034171343\\
7	0.728211033691746\\
8	1.22290640977326\\
9	1.6381739378461\\
10	1.97989786975532\\
};
\addlegendentry{Last value predictor}

\addplot [color=black, line width=1.0pt]
  table[row sep=crcr]{%
1	-0.701207302483621\\
2	-0.701283172070688\\
3	-0.705591436315899\\
4	-0.711898244289577\\
5	-0.706736370785316\\
6	-0.709260539319819\\
7	-0.713514136413639\\
8	-0.711808915548028\\
9	-0.716174447578272\\
10	-0.702238800994987\\
};
\addlegendentry{Baseline}

\end{axis}

\begin{axis}[%
width=3.382in,
height=3.5in,
at={(9in,0.573in)},
xmin=1,
xmax=10,
ymin=-8,
ymax=3,
axis background/.style={fill=white},
title style={font=\bfseries},
title={(c) Setup 3},
axis x line*=bottom,
axis y line*=left,
xmajorgrids,
ymajorgrids
]
\node[below right, align=left]
at (rel axis cs:.05,0.96) { $\eta = 5$ slots\\($\nu = 0.0765$ m/slot)};
\addplot [color=black, line width=1.0pt, forget plot]
  table[row sep=crcr]{%
1	-0.739470366024188\\
2	-0.744916330478484\\
3	-0.739106374428443\\
4	-0.737713453739772\\
5	-0.740539343139986\\
6	-0.738336193709358\\
8	-0.730229435806386\\
9	-0.732740563697719\\
10	-0.732627121957281\\
};
\addplot [color=blue, line width=1.0pt, mark size=4.0pt, mark=o, mark options={solid, blue}, forget plot]
  table[row sep=crcr]{%
1	-6.4741109786371\\
2	-4.93002293324816\\
3	-3.25573797328844\\
4	-2.57207005376096\\
5	-1.73925939238625\\
6	-1.40842389502357\\
7	-1.01848154365968\\
8	-0.759202248724856\\
9	-0.513823378068247\\
10	-0.498506768080635\\
};
\addplot [color=mycolor1, line width=1.0pt, mark size=4.0pt, mark=+, mark options={solid, mycolor1}, forget plot]
  table[row sep=crcr]{%
1	-7.72695912453207\\
2	-4.81850439994405\\
3	-3.13449379597263\\
4	-1.91864770219505\\
5	-0.996851093212232\\
6	-0.258401178495115\\
7	0.397386941572819\\
8	0.923887366536576\\
9	1.38091281820479\\
10	1.80562722848884\\
};
\addplot [color=red, line width=1.0pt, mark size=4.0pt, mark=x, mark options={solid, red}, forget plot]
  table[row sep=crcr]{%
1	-7.48614249328476\\
2	-4.64244831937591\\
3	-3.10597484811562\\
4	-1.97183206372463\\
5	-1.0400627488286\\
6	-0.25998591793433\\
7	0.40869610261236\\
8	0.892276681612071\\
9	1.29265182445719\\
10	1.7215360228314\\
};
\end{axis}

\begin{axis}[%
width=3.382in,
height=3.5in,
at={(2in,0.573in)},
xmin=1,
xmax=10,
ymin=-8,
ymax=3,
ylabel style={font=\bfseries\color{white!15!black}},
ylabel={NMSE in dB},
axis background/.style={fill=white},
title style={font=\bfseries},
title={(a) Setup 1},
axis x line*=bottom,
axis y line*=left,
xmajorgrids,
ymajorgrids
]
\node[below right, align=left]
at (rel axis cs:.05,0.96) { $\eta = 50$ slots\\($\nu = 0.0077$ m/slot)};
\addplot [color=mycolor1, line width=1.0pt, mark size=4.0pt, mark=+, mark options={solid, mycolor1}, forget plot]
  table[row sep=crcr]{%
1	-5.90905946775689\\
2	-2.22269925863267\\
3	-0.349530501210394\\
4	0.659322564373211\\
5	1.14356045408234\\
6	1.38233465639766\\
7	1.52903640936282\\
8	1.72052455332925\\
9	1.96254711580976\\
10	2.20727689189946\\
};
\addplot [color=red, line width=1.0pt, mark size=4.0pt, mark=x, mark options={solid, red}, forget plot]
  table[row sep=crcr]{%
1	-5.88018620395523\\
2	-2.55577936795256\\
3	-1.29882148834501\\
4	-0.790437593566242\\
5	-0.616224342785241\\
6	-0.55473377781429\\
7	-0.646651793150804\\
8	-0.623775337519175\\
9	-0.562990217997243\\
10	-0.522683973275699\\
};
\addplot [color=blue, line width=1.0pt, mark size=4.0pt, mark=o, mark options={solid, blue}, forget plot]
  table[row sep=crcr]{%
1	-6.24785813632853\\
2	-3.04441314048043\\
3	-1.73186802784623\\
4	-1.06202387250358\\
5	-0.852285852632027\\
6	-0.752409259863272\\
7	-0.70026217118342\\
8	-0.641612639555493\\
9	-0.567123645232977\\
10	-0.547609873363349\\
};
\addplot [color=black, line width=1.0pt, forget plot]
  table[row sep=crcr]{%
1	-0.589827636925596\\
2	-0.588682436865499\\
3	-0.597994208813692\\
4	-0.594817304141451\\
5	-0.594397245672333\\
6	-0.590407568874241\\
7	-0.593305667670789\\
8	-0.591986267250862\\
9	-0.588305554667219\\
10	-0.585884523055721\\
};
\end{axis}

\end{tikzpicture}%
 }
\caption{Predictor evaluation for increasing node mobility. Comparison between the proposed interference predictor (blue/o), a channel predictor that ignores the traffic influence (red/x), and a predictor that outputs as prediction its current observation (green/+). The interference coherence time for $\theta = 0.25$ is $5$ time slots for subfigures a and b, and $3$ time slots for subfigure c. The baseline of using the mean value of the interference as prediction is shown for reference. } 
\label{Fig:Predictor2}
\end{figure*}
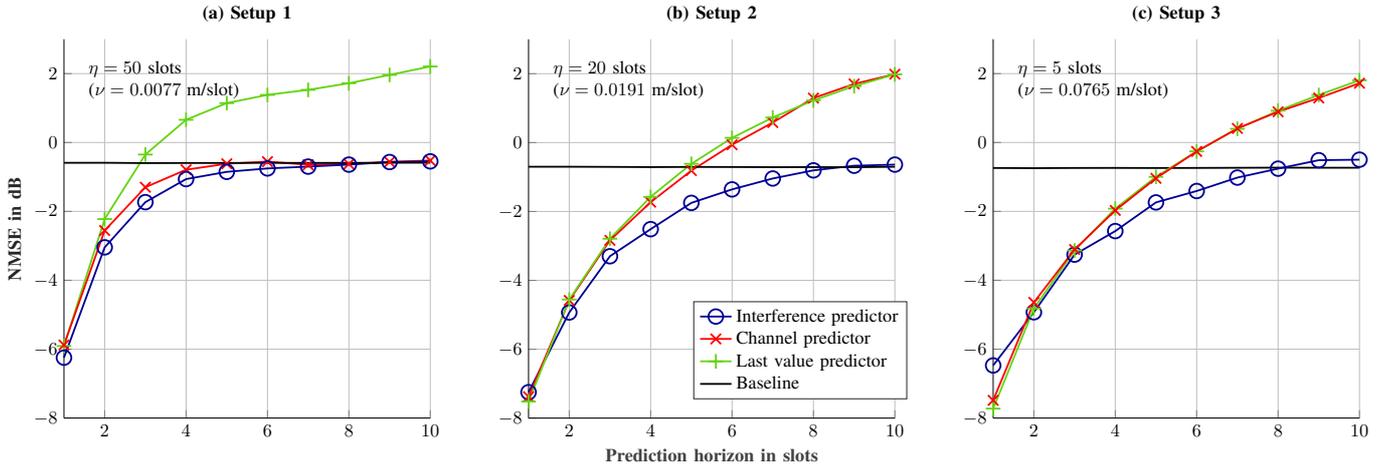  

An interference predictor can be constructed from the ARMA($p,q$) approximation of the interference process. Expressing the model coefficients as polynomials in the lag operator $L$ and provided that the roots of the polynomial on $a_n$ are outside the unit circle, the interference sequence becomes the filtered output of the noise sequence $\epsilon(t)$:
\begin{equation}
	i(t) = \frac{b_0 + b_1L +\ldots + b_qL^q}{a_0 + a_1L +\ldots + a_pL^p}\:\epsilon(t).
	\label{eq:ARMAfilter}
\end{equation}

This filter form suggests that it is possible to predict interference samples $\hat{i}(t+\Delta)$ by feeding \eqref{eq:ARMAfilter} with a white noise sequence. However, such open loop prediction has poor accuracy given the sample autocorrelation dispersion between different realizations of the interference process. Nevertheless, a closed loop Kalman formulation of the filtering problem gives accurate predictions as will be shown in the next section.

For deriving the Kalman filter recursion, we first map our ARMA model into a state space form:
\begin{eqnarray}
\label{eq:Kalman_1}
 {\bf x}(t+1) &=& {\bf A}{\bf x}(t)+{\bf B}\epsilon(t) \\
\label{eq:Kalman_Exit}
      i(t)    &=& {\bf C}{\bf x}(t),
\end{eqnarray}
with state vector {\bf x} of size $p\times 1$, transition matrix 
\begin{equation}
	\label{eq:Kalman_A}
	{\bf A} = 
	\begin{bmatrix}
	a_1    &a_2    &\cdots &a_{p-1} &a_p   \\ 
  1      &0      &\cdots &0       &0     \\ 
  0      &1      &\cdots &0       &0     \\ 
  \vdots &\vdots &\ddots &\vdots  &\vdots\\ 
  0      &0      &\cdots &1       &0
	\end{bmatrix},
\end{equation}
 and ${\bf B}=[1\ 0 \cdots 0]^{\rm T}$ and ${\bf C}=[0 \cdots 0\ b_0 \cdots b_q]$ of length $p$.

Assuming unitary process and measurement noise and following \cite{Chui_Chen}, the Kalman recursion for tracking $i(t)$ can be initialized with an all-zero state vector and an initial error covariance ${\bf P} = {\bf B}{\bf B}^{\rm T}$. At each iteration of the filter, the measurement update is 
\begin{eqnarray}
\label{eq:Kalman_Meas}
	{\bf M} &=& {\bf P}{\bf C}^{\rm T}\left({\bf C}{\bf P}{\bf C}^{\rm T}+1\right)^{-1}\\
  {\bf x}(t) &=& {\bf x}(t) + {\bf M}(i(t)-{\bf C}{\bf x}(t)) \\
   \label{eq:Kalman_Meas_end}
   {\bf P} &=& ({\bf I}_p-{\bf M}{\bf C}){\bf P},
\end{eqnarray}
with the innovation gain ${\bf M}$ and identity matrix ${\bf I}_p$ of size $p$. The time update is
\begin{eqnarray}
\label{eq:Kalman_Update}
  {\bf x}(t+1) &=& {\bf A}{\bf x}(t) + {\bf B}(i(t)-{\bf C}{\bf x}(t)) \\
	\label{eq:Kalman_Update2}
   {\bf P} &=& {\bf A}{\bf P}{\bf A}^{\rm T}+ {\bf B}{\bf B}^{\rm T}.
\end{eqnarray}

As the interference process is time invariant and the model described by \eqref{eq:ARMAfilter} and \eqref{eq:Kalman_1} is assumed to be reachable and detectable, 
the error covariance ${\bf P}$ converges to a constant~\cite{Chui_Chen}, and \eqref{eq:Kalman_Meas} to \eqref{eq:Kalman_Update2} reduce to a steady-state Kalman filter of gain~${\bf K}$:
\begin{equation}
\label{eq:Kalman_SS}
  {\bf x}(t+1) = {\bf A}{\bf x}(t) + {\bf K}(i(t)-{\bf C}{\bf x}(t)).
\end{equation}

The interference prediction for a prediction horizon $\Delta$ can finally be found by extrapolating the state vector $\Delta$ time slots into the future on each iteration and using~\eqref{eq:Kalman_Exit}:
\begin{equation}
\label{eq:Kalman_Predict}
  \tilde{\bf x} = {\bf A}^{\Delta-1}{\bf x}(t+1), \qquad  \hat{i}(t+\Delta) = {\bf C}\tilde{\bf x}.
\end{equation} 
The gain ${\bf K}$ can be computed offline, which reduces the online effort of the predictor to the state update in \eqref{eq:Kalman_SS} and to the extrapolation of the filter through \eqref{eq:Kalman_Predict}.

Let us summarize the steps with the block diagram in Fig.~\ref{Fig:BlockDiagram}. From a network model abstraction, the interference autocorrelation is obtained from \eqref{eq:AutocorrFunction}. This is used to parameterize an ARMA($p,q$) model, which is inserted into a steady-state Kalman filter to complete the predictor's offline design. Predicted  samples $\hat{i}(t+\Delta)$ are filtered samples of~$i(t)$.

\subsection{Complexity Analysis}
We quantify the implementation cost in terms of  memory requirements and the number of sum and product operations needed to compute one predicted sample. These follow from \eqref{eq:Kalman_SS} for the proposed steady-state Kalman recursion and from \eqref{eq:Kalman_Meas} to \eqref{eq:Kalman_Update2} for the standard Kalman predictor. The prediction step of \eqref{eq:Kalman_Predict} completes the analysis for both. The implementation cost of the standard Kalman recursion helps to assess the computational gains from the steady-state formulation. 

\subsubsection{Memory}
For a model of order $p$, the standard Kalman filter stores the matrices $\bf A$ and $\bf P$, vectors $\bf B$, $\bf C$, $\bf M$, $\bf x$, and $\tilde{\bf x}$, and scalars $i(t)$ and $\hat{i}(t+\Delta)$, requiring $2p^2+5p+2$ floating point memory units. The steady-state formulation requires storage for matrix $\bf A$, vectors $\bf C$, $\bf K$, $\bf x$, and $\tilde{\bf x}$, and scalars $i(t)$ and $\hat{i}(t+\Delta)$, amounting to $p^2+4p+2$ memory units. For practical model orders (Fig.~\ref{Fig:ApproxError}), both formulations have low and similar memory requirements. 

\subsubsection{Computations}
We make some approximations that simplify  analysis but still capture the main floating point operation costs. The vector-vector product is approximated to require $2p$ operations, the vector-matrix product $2p^2$ operations, and the matrix-matrix product $2p^3$ operations. Operations with scalars are disregarded. 
We distinguish between computations for the measurement and time updates, and for the prediction. We find that a standard Kalman predictor evaluates \eqref{eq:Kalman_Meas} to \eqref{eq:Kalman_Meas_end} for the measurement update, and \eqref{eq:Kalman_Update} and \eqref{eq:Kalman_Update2} for the time update. In the case of the steady-state Kalman predictor, the measurement update is not evaluated since the steady-state gain $\bf K$ is computed offline. For the time update, \eqref{eq:Kalman_SS} is evaluated. Finally, both predictors use \eqref{eq:Kalman_Predict} to obtain the predicted interference sample. Table~\ref{tab:Complexity} summarizes the computational costs. 

\begin{table}[th]
\begin{center}
\caption{Implementation Cost in Operations Required}
\label{tab:Complexity}
\renewcommand{\arraystretch}{1.3}
\begin{tabular}{lcc}
\hline
Type of operation & Standard predictor & Steady-state predictor\\
\hline
{Measurement update} & $2p^3+4p^2+9p$ & --\\
{Time update}  & $4p^3+2p^2+6p$ & $p^2+3p$ \\
{Prediction}   & \multicolumn{2}{c}{$2(\Delta-2)p^3+2p^2+2p$}\\ \hline
\end{tabular}
\end{center}
\end{table}

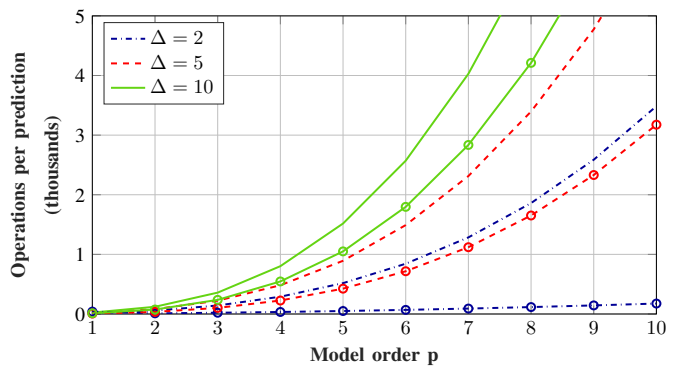
\begin{figure}[htbp]
 \centering
  \resizebox{.49\textwidth}{!}{
  \definecolor{mycolor1}{rgb}{0.39216,0.83137,0.07451}%
\begin{tikzpicture}

\begin{axis}[%
width=4.521in,
height=2.7in,
at={(0.894in,0.467in)},
xmin=1,
xmax=10,
xlabel style={font=\bfseries\color{white!15!black}},
xlabel={Model order p},
ymin=0,
ymax=10,
ytick={0,2,4,6,8,10},
yticklabels={{0},{1},{2},{3},{4},{5},{6},{7},{8},{9},{10}},
ylabel style={font=\bfseries\color{white!15!black}, align=center},
ylabel={Operations per prediction\\[1ex]           (thousands)},
axis background/.style={fill=white},
xmajorgrids,
ymajorgrids,
legend style={at={(0.02,0.703)}, anchor=south west, legend cell align=left, align=left, draw=white!15!black}
]
\addplot [color=blue, dashdotted, line width=1.0pt]
  table[row sep=crcr]{%
1	.031\\
2	.114\\
3	.285\\
4	.580\\
5	1.035\\
6	1.686\\
7	2.569\\
8	3.720\\
9	5.175\\
10	6.970\\
};
\addlegendentry{$\Delta = 2$}

\addplot [color=red, dashed, line width=1.0pt]
  table[row sep=crcr]{%
1	.037\\
2	.162\\
3	.447\\
4	.964\\
5	1.785\\
6	2.982\\
7	4.627\\
8	6.792\\
9	9.549\\
9.42414498684593	11.000\\
};
\addlegendentry{$\Delta = 5$}

\addplot [color=mycolor1, line width=1.0pt]
  table[row sep=crcr]{%
1	.047\\
2	.242\\
3	.717\\
4	1.604\\
5	3.035\\
6	5.142\\
7	8.057\\
7.76342412451413	11.000\\
};
\addlegendentry{$\Delta = 10$}

\addplot [color=blue, dashdotted, line width=1.0pt, mark=o, mark options={solid, blue}, forget plot]
  table[row sep=crcr]{%
1	.08\\
2	.022\\
3	.042\\
4	.068\\
5	.100\\
6	.138\\
7	.182\\
8	.232\\
9	.288\\
10	.350\\
};
\addplot [color=red, dashed, line width=1.0pt, mark=o, mark options={solid, red}, forget plot]
  table[row sep=crcr]{%
1	.014\\
2	.070\\
3	.204\\
4	.452\\
5	.850\\
6	1.434\\
7	2.240\\
8	3.304\\
9	4.662\\
10	6.350\\
};
\addplot [color=mycolor1, line width=1.0pt, mark=o, mark options={solid, mycolor1}, forget plot]
  table[row sep=crcr]{%
1	.024\\
2	.150\\
3	.474\\
4	1.092\\
5	2.100\\
6	3.594\\
7	5.670\\
8	8.424\\
8.73015873015902	11.000\\
};
\end{axis}

\end{tikzpicture}%

 }
\caption{Predictor complexity in terms of the number of floating point operations required per predicted sample. Results are shown for increasing model orders $p$ with prediction horizons of $\Delta=10$ (solid green), $5$ (dashed red), and $2$ (dash-dotted blue). Traces with (without) marker correspond to the steady-state (standard) Kalman filter predictor. }
\label{Fig:Complexity}
\end{figure}

Figure~\ref{Fig:Complexity} shows the advantage of the steady-state formulation in terms of complexity. The computational cost for a prediction horizon of $\Delta=5$ using the steady-state predictor (dashed trace with marker) is lower than that of $\Delta=2$ using the standard Kalman predictor (dashed-dotted trace), for all~$p$. 
The complexity savings from the steady-state formulation are more significant for shorter prediction horizons. For $p=7$, the steady-state Kalman predictor requires 52\% fewer operations for $\Delta=5$ and 29\% fewer for $\Delta=10$. As the model order increases, the savings of the steady-state formulation slightly decrease for a given $\Delta$. Both predictors show low implementation cost, with the steady-state version having the lowest in all cases. In summary, for implementation on highly constrained devices\,---\,like those employed in wireless sensor networks\,---\,the proposed steady-state predictor is convenient. 

\section{Evaluation of \mbox{Interference Prediction}}
\label{SEC:Evaluation} 
The prediction performance is evaluated in terms of the normalized mean square error
\begin{equation}
\label{eq:NMSE_def}
  {\rm NMSE} = \frac{\sum_t(i(t+\Delta)-\hat{i}(t+\Delta))^2}{\sum_t i(t+\Delta)^2},
\end{equation}
which is a commonly used metric in channel prediction (see \cite{5951808, Hallen_Dec07, somethesis}), as the normalization with the squared power of the input signal allows to average realizations with different powers. Error results are always averaged over 10,000 network realizations with 1,000 time slots interference sequences each.

\subsection{Evaluation in the Base System}

As a first step, the interference predictor is applied in the base system that was used for its parameterization: PPP, Rayleigh fading, and fixed message lengths. The setups from Table~\ref{tab:Scenarios} are used with a node density of $\lambda = 0.01$ over an area of $10,\!000$ square units, a path loss exponent $\alpha = 3$, and all nodes using unitary transmit power $\kappa=1$. 

Figure~\ref{Fig:Predictor2} plots the NMSE over the prediction horizon in slots. Three predictors are compared: the interference predictor, a channel predictor, and a last value predictor. The channel predictor uses the same methodology as the interference predictor but with $\rho(\tau)$ accounting for the channel as the sole source of correlation ($\txp=1$), and illustrates the impact of ignoring the traffic. The last value predictor simply takes its current interference observation as prediction. It is a performance reference for determining whether the extra (but low) complexity of the proposed predictor is justified. A ``mean value predictor'' serves as a baseline; it completely disregards interference dynamics and uses the mean value of the interference as prediction. As expected, the NMSE increases with an increasing prediction horizon for all three predictors in all setups. The interference predictor outperforms both the channel  and last value predictors. Mobility increases from subplots (a) to (c) (i.e., Setups 1 to 3 from Table~\ref{tab:Scenarios}). The channel predictor gets close to the interference predictor for the low mobility scenario (a). For higher node speeds, the channel predictor quickly degrades to the level of the last value predictor. The interference predictor crosses the baseline in all cases for a prediction horizon of eight slots. This horizon is significantly larger than the three and five slots obtained for the last value predictor (Subfigure~(a) and Subfigures~(b) and (c), respectively). This may justify its implementation cost. 

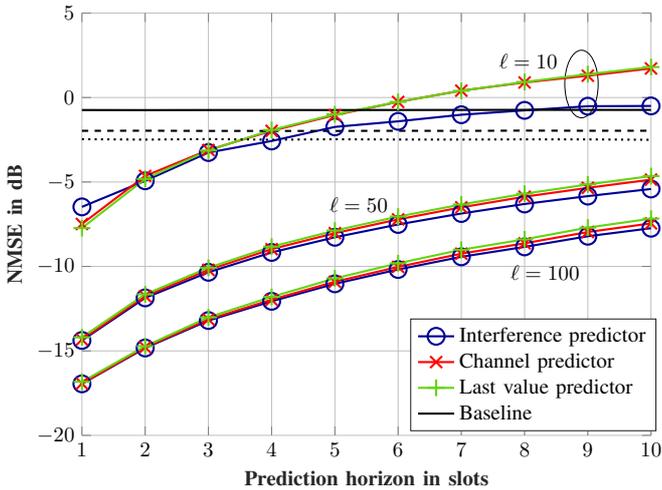
\begin{figure}[htbp]
 \centering
  \resizebox{.49\textwidth}{!}{
  \definecolor{mycolor1}{rgb}{0.39216,0.83137,0.07451}%
\begin{tikzpicture}

\begin{axis}[%
width=4.347in,
height=3.409in,
at={(0.829in,0.529in)},
xmin=1,
xmax=10,
xlabel style={font=\bfseries\color{white!15!black}},
xlabel={Prediction horizon in slots},
ymin=-20,
ymax=5,
ylabel style={font=\bfseries\color{white!15!black}},
ylabel={NMSE in dB},
axis background/.style={fill=white},
xmajorgrids,
ymajorgrids,
axis line style={draw=none},
legend style={at={(0.577,0.01)}, anchor=south west, legend cell align=left, align=left, draw=white!10!black}
]
\draw [black] (axis cs:8.9,0.8) ellipse [x radius=0.264286, y radius=2];
\node[below right, align=left, font=\bfseries]
at (rel axis cs:0.422,0.58) {$\ell = 50$};
\node[below right, align=left, font=\bfseries]
at (rel axis cs:0.74,0.42) {$\ell = 100$};
\node[below right, align=left, font=\bfseries]
at (rel axis cs:0.72,0.92) {$\ell = 10$};
\addplot [color=blue, line width=1.0pt, mark size=4.0pt, mark=o, mark options={solid, blue}]
  table[row sep=crcr]{%
1	-6.4741109786371\\
2	-4.93002293324816\\
3	-3.25573797328844\\
4	-2.57207005376096\\
5	-1.73925939238625\\
6	-1.40842389502357\\
7	-1.01848154365968\\
8	-0.759202248724856\\
9	-0.513823378068247\\
10	-0.498506768080635\\
};
\addlegendentry{Interference predictor}

\addplot [color=red, line width=1.0pt, mark size=4.0pt, mark=x, mark options={solid, red}]
  table[row sep=crcr]{%
1	-7.48614249328476\\
2	-4.64244831937591\\
3	-3.10597484811562\\
4	-1.97183206372463\\
5	-1.0400627488286\\
6	-0.25998591793433\\
7	0.40869610261236\\
8	0.892276681612071\\
9	1.29265182445719\\
10	1.7215360228314\\
};
\addlegendentry{Channel predictor}

\addplot [color=mycolor1, line width=1.0pt, mark size=4.0pt, mark=+, mark options={solid, mycolor1}]
  table[row sep=crcr]{%
1	-7.72695912453207\\
2	-4.81850439994405\\
3	-3.13449379597263\\
4	-1.91864770219505\\
5	-0.996851093212232\\
6	-0.258401178495115\\
7	0.397386941572819\\
8	0.923887366536576\\
9	1.38091281820479\\
10	1.80562722848884\\
};
\addlegendentry{Last value predictor}

\addplot [color=black, line width=1.0pt]
  table[row sep=crcr]{%
1	-0.739470366024188\\
2	-0.744916330478484\\
3	-0.739106374428443\\
4	-0.737713453739772\\
5	-0.740539343139986\\
7	-0.734681045486601\\
8	-0.730229435806386\\
10	-0.732627121957281\\
};
\addlegendentry{Baseline}

\addplot [color=blue, line width=1.0pt, mark size=4.0pt, mark=o, mark options={solid, blue}, forget plot]
  table[row sep=crcr]{%
1	-14.3809831728194\\
2	-11.8608897713134\\
3	-10.3464224944265\\
4	-9.17054061152743\\
5	-8.27411657480581\\
6	-7.51519262159298\\
7	-6.86626203995195\\
8	-6.30173693191844\\
9	-5.83597132450511\\
10	-5.41436759499926\\
};
\addplot [color=red, line width=1.0pt, mark size=4.0pt, mark=x, mark options={solid, red}, forget plot]
  table[row sep=crcr]{%
1	-14.3284817940719\\
2	-11.7608081778569\\
3	-10.2008459151783\\
4	-8.97636201050121\\
5	-8.03033898994696\\
6	-7.21381981618221\\
7	-6.50162238289808\\
8	-5.8718172601584\\
9	-5.34711905442878\\
10	-4.85959432581463\\
};
\addplot [color=mycolor1, line width=1.0pt, mark size=4.0pt, mark=+, mark options={solid, mycolor1}, forget plot]
  table[row sep=crcr]{%
1	-14.19371117596\\
2	-11.6516543362141\\
3	-10.0773850481377\\
4	-8.8303108427567\\
5	-7.87593922792763\\
6	-7.04771122581954\\
7	-6.33288544561734\\
8	-5.69257582230857\\
9	-5.14523466580284\\
10	-4.64909681275517\\
};
\addplot [color=black, dashed, line width=1.0pt, forget plot]
  table[row sep=crcr]{%
1	-1.96987639282632\\
2	-1.96375897454121\\
3	-1.9768794152324\\
5	-1.97755689651888\\
6	-1.96866045768167\\
7	-1.97457679687713\\
8	-1.95719492126987\\
9	-1.96660697651065\\
10	-1.96544699807237\\
};
\addplot [color=blue, line width=1.0pt, mark size=4.0pt, mark=o, mark options={solid, blue}, forget plot]
  table[row sep=crcr]{%
1	-16.9505258812088\\
2	-14.8275225406057\\
3	-13.2017969750983\\
4	-12.0629224138892\\
5	-11.0288496962969\\
6	-10.1808981796212\\
7	-9.43047762454571\\
8	-8.84334414949249\\
9	-8.21255582282158\\
10	-7.73772796376376\\
};
\addplot [color=black, dotted, line width=1.0pt, forget plot]
  table[row sep=crcr]{%
1	-2.46531005056788\\
2	-2.49203213413267\\
3	-2.4578947383891\\
4	-2.49711480082319\\
5	-2.47067346975496\\
6	-2.48610352885941\\
7	-2.48895905553187\\
8	-2.47660620905703\\
9	-2.47812047650773\\
10	-2.50206828310866\\
};
\addplot [color=red, line width=1.0pt, mark size=4.0pt, mark=x, mark options={solid, red}, forget plot]
  table[row sep=crcr]{%
1	-16.9323010974673\\
2	-14.7863460470771\\
3	-13.133335084021\\
4	-11.9693408558805\\
5	-10.9073985696747\\
6	-10.0273138435277\\
7	-9.24470256518732\\
8	-8.63058999578735\\
9	-7.9633873190061\\
10	-7.45106298533726\\
};
\addplot [color=mycolor1, line width=1.0pt, mark size=4.0pt, mark=+, mark options={solid, mycolor1}, forget plot]
  table[row sep=crcr]{%
1	-16.8298285298168\\
2	-14.7206319676239\\
3	-13.0103542965487\\
4	-11.80367021201\\
5	-10.7075881295952\\
6	-9.81454623772319\\
7	-9.01197650596737\\
8	-8.39615608099238\\
9	-7.70121463912194\\
10	-7.17069226569767\\
};
\end{axis}
\end{tikzpicture}%
 }
\caption{Evaluation of the predictors for Setup~1 with different message lengths. The interference coherence time for $\theta = 0.25$ is $5$, $21$, and $47$ slots for $\ell = 10$, $50$, and $100$, respectively.
The mean value predictor baseline is shown for $\ell = 10$, $50$, and $100$.} 
\label{Fig:PredictorMsgs}
\end{figure}  

Figure~\ref{Fig:PredictorMsgs} explores the predictors' behavior (adjusted for Setup~1 from Table~\ref{tab:Scenarios}) as the message length changes: Longer messages result in more correlated interference and thus lower prediction error. In all cases, the interference predictor outperforms the others; however, the longer the messages, the more similar the predictors perform. For $\ell=10$, the interference predictor attains horizons of eight slots with an error below the baseline, whereas the last value and channel predictors cross above the baseline for an horizon of five slots. 

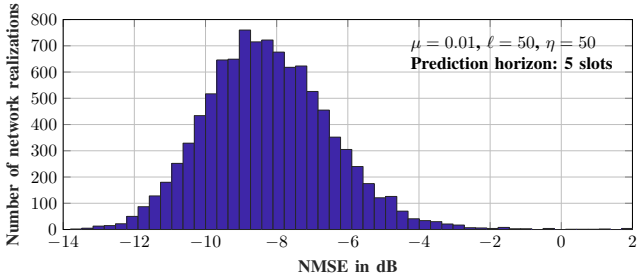
\begin{figure}[htbp]
 \centering
  \resizebox{.47\textwidth}{!}{
  \begin{tikzpicture}

\begin{axis}[%
width=5.133in,
height=2.3in,
at={(0.615in,0.431in)},
point meta min=1,
point meta max=2,
colormap={mymap}{[1pt] rgb(0pt)=(0.2422,0.1504,0.6603); rgb(1pt)=(0.2444,0.1534,0.6728); rgb(2pt)=(0.2464,0.1569,0.6847); rgb(3pt)=(0.2484,0.1607,0.6961); rgb(4pt)=(0.2503,0.1648,0.7071); rgb(5pt)=(0.2522,0.1689,0.7179); rgb(6pt)=(0.254,0.1732,0.7286); rgb(7pt)=(0.2558,0.1773,0.7393); rgb(8pt)=(0.2576,0.1814,0.7501); rgb(9pt)=(0.2594,0.1854,0.761); rgb(11pt)=(0.2628,0.1932,0.7828); rgb(12pt)=(0.2645,0.1972,0.7937); rgb(13pt)=(0.2661,0.2011,0.8043); rgb(14pt)=(0.2676,0.2052,0.8148); rgb(15pt)=(0.2691,0.2094,0.8249); rgb(16pt)=(0.2704,0.2138,0.8346); rgb(17pt)=(0.2717,0.2184,0.8439); rgb(18pt)=(0.2729,0.2231,0.8528); rgb(19pt)=(0.274,0.228,0.8612); rgb(20pt)=(0.2749,0.233,0.8692); rgb(21pt)=(0.2758,0.2382,0.8767); rgb(22pt)=(0.2766,0.2435,0.884); rgb(23pt)=(0.2774,0.2489,0.8908); rgb(24pt)=(0.2781,0.2543,0.8973); rgb(25pt)=(0.2788,0.2598,0.9035); rgb(26pt)=(0.2794,0.2653,0.9094); rgb(27pt)=(0.2798,0.2708,0.915); rgb(28pt)=(0.2802,0.2764,0.9204); rgb(29pt)=(0.2806,0.2819,0.9255); rgb(30pt)=(0.2809,0.2875,0.9305); rgb(31pt)=(0.2811,0.293,0.9352); rgb(32pt)=(0.2813,0.2985,0.9397); rgb(33pt)=(0.2814,0.304,0.9441); rgb(34pt)=(0.2814,0.3095,0.9483); rgb(35pt)=(0.2813,0.315,0.9524); rgb(36pt)=(0.2811,0.3204,0.9563); rgb(37pt)=(0.2809,0.3259,0.96); rgb(38pt)=(0.2807,0.3313,0.9636); rgb(39pt)=(0.2803,0.3367,0.967); rgb(40pt)=(0.2798,0.3421,0.9702); rgb(41pt)=(0.2791,0.3475,0.9733); rgb(42pt)=(0.2784,0.3529,0.9763); rgb(43pt)=(0.2776,0.3583,0.9791); rgb(44pt)=(0.2766,0.3638,0.9817); rgb(45pt)=(0.2754,0.3693,0.984); rgb(46pt)=(0.2741,0.3748,0.9862); rgb(47pt)=(0.2726,0.3804,0.9881); rgb(48pt)=(0.271,0.386,0.9898); rgb(49pt)=(0.2691,0.3916,0.9912); rgb(50pt)=(0.267,0.3973,0.9924); rgb(51pt)=(0.2647,0.403,0.9935); rgb(52pt)=(0.2621,0.4088,0.9946); rgb(53pt)=(0.2591,0.4145,0.9955); rgb(54pt)=(0.2556,0.4203,0.9965); rgb(55pt)=(0.2517,0.4261,0.9974); rgb(56pt)=(0.2473,0.4319,0.9983); rgb(57pt)=(0.2424,0.4378,0.9991); rgb(58pt)=(0.2369,0.4437,0.9996); rgb(59pt)=(0.2311,0.4497,0.9995); rgb(60pt)=(0.225,0.4559,0.9985); rgb(61pt)=(0.2189,0.462,0.9968); rgb(62pt)=(0.2128,0.4682,0.9948); rgb(63pt)=(0.2066,0.4743,0.9926); rgb(64pt)=(0.2006,0.4803,0.9906); rgb(65pt)=(0.195,0.4861,0.9887); rgb(66pt)=(0.1903,0.4919,0.9867); rgb(67pt)=(0.1869,0.4975,0.9844); rgb(68pt)=(0.1847,0.503,0.9819); rgb(69pt)=(0.1831,0.5084,0.9793); rgb(70pt)=(0.1818,0.5138,0.9766); rgb(71pt)=(0.1806,0.5191,0.9738); rgb(72pt)=(0.1795,0.5244,0.9709); rgb(73pt)=(0.1785,0.5296,0.9677); rgb(74pt)=(0.1778,0.5349,0.9641); rgb(75pt)=(0.1773,0.5401,0.9602); rgb(76pt)=(0.1768,0.5452,0.956); rgb(77pt)=(0.1764,0.5504,0.9516); rgb(78pt)=(0.1755,0.5554,0.9473); rgb(79pt)=(0.174,0.5605,0.9432); rgb(80pt)=(0.1716,0.5655,0.9393); rgb(81pt)=(0.1686,0.5705,0.9357); rgb(82pt)=(0.1649,0.5755,0.9323); rgb(83pt)=(0.161,0.5805,0.9289); rgb(84pt)=(0.1573,0.5854,0.9254); rgb(85pt)=(0.154,0.5902,0.9218); rgb(86pt)=(0.1513,0.595,0.9182); rgb(87pt)=(0.1492,0.5997,0.9147); rgb(88pt)=(0.1475,0.6043,0.9113); rgb(89pt)=(0.1461,0.6089,0.908); rgb(90pt)=(0.1446,0.6135,0.905); rgb(91pt)=(0.1429,0.618,0.9022); rgb(92pt)=(0.1408,0.6226,0.8998); rgb(93pt)=(0.1383,0.6272,0.8975); rgb(94pt)=(0.1354,0.6317,0.8953); rgb(95pt)=(0.1321,0.6363,0.8932); rgb(96pt)=(0.1288,0.6408,0.891); rgb(97pt)=(0.1253,0.6453,0.8887); rgb(98pt)=(0.1219,0.6497,0.8862); rgb(99pt)=(0.1185,0.6541,0.8834); rgb(100pt)=(0.1152,0.6584,0.8804); rgb(101pt)=(0.1119,0.6627,0.877); rgb(102pt)=(0.1085,0.6669,0.8734); rgb(103pt)=(0.1048,0.671,0.8695); rgb(104pt)=(0.1009,0.675,0.8653); rgb(105pt)=(0.0964,0.6789,0.8609); rgb(106pt)=(0.0914,0.6828,0.8562); rgb(107pt)=(0.0855,0.6865,0.8513); rgb(108pt)=(0.0789,0.6902,0.8462); rgb(109pt)=(0.0713,0.6938,0.8409); rgb(110pt)=(0.0628,0.6972,0.8355); rgb(111pt)=(0.0535,0.7006,0.8299); rgb(112pt)=(0.0433,0.7039,0.8242); rgb(113pt)=(0.0328,0.7071,0.8183); rgb(114pt)=(0.0234,0.7103,0.8124); rgb(115pt)=(0.0155,0.7133,0.8064); rgb(116pt)=(0.0091,0.7163,0.8003); rgb(117pt)=(0.0046,0.7192,0.7941); rgb(118pt)=(0.0019,0.722,0.7878); rgb(119pt)=(0.0009,0.7248,0.7815); rgb(120pt)=(0.0018,0.7275,0.7752); rgb(121pt)=(0.0046,0.7301,0.7688); rgb(122pt)=(0.0094,0.7327,0.7623); rgb(123pt)=(0.0162,0.7352,0.7558); rgb(124pt)=(0.0253,0.7376,0.7492); rgb(125pt)=(0.0369,0.74,0.7426); rgb(126pt)=(0.0504,0.7423,0.7359); rgb(127pt)=(0.0638,0.7446,0.7292); rgb(128pt)=(0.077,0.7468,0.7224); rgb(129pt)=(0.0899,0.7489,0.7156); rgb(130pt)=(0.1023,0.751,0.7088); rgb(131pt)=(0.1141,0.7531,0.7019); rgb(132pt)=(0.1252,0.7552,0.695); rgb(133pt)=(0.1354,0.7572,0.6881); rgb(134pt)=(0.1448,0.7593,0.6812); rgb(135pt)=(0.1532,0.7614,0.6741); rgb(136pt)=(0.1609,0.7635,0.6671); rgb(137pt)=(0.1678,0.7656,0.6599); rgb(138pt)=(0.1741,0.7678,0.6527); rgb(139pt)=(0.1799,0.7699,0.6454); rgb(140pt)=(0.1853,0.7721,0.6379); rgb(141pt)=(0.1905,0.7743,0.6303); rgb(142pt)=(0.1954,0.7765,0.6225); rgb(143pt)=(0.2003,0.7787,0.6146); rgb(144pt)=(0.2061,0.7808,0.6065); rgb(145pt)=(0.2118,0.7828,0.5983); rgb(146pt)=(0.2178,0.7849,0.5899); rgb(147pt)=(0.2244,0.7869,0.5813); rgb(148pt)=(0.2318,0.7887,0.5725); rgb(149pt)=(0.2401,0.7905,0.5636); rgb(150pt)=(0.2491,0.7922,0.5546); rgb(151pt)=(0.2589,0.7937,0.5454); rgb(152pt)=(0.2695,0.7951,0.536); rgb(153pt)=(0.2809,0.7964,0.5266); rgb(154pt)=(0.2929,0.7975,0.517); rgb(155pt)=(0.3052,0.7985,0.5074); rgb(156pt)=(0.3176,0.7994,0.4975); rgb(157pt)=(0.3301,0.8002,0.4876); rgb(158pt)=(0.3424,0.8009,0.4774); rgb(159pt)=(0.3548,0.8016,0.4669); rgb(160pt)=(0.3671,0.8021,0.4563); rgb(161pt)=(0.3795,0.8026,0.4454); rgb(162pt)=(0.3921,0.8029,0.4344); rgb(163pt)=(0.405,0.8031,0.4233); rgb(164pt)=(0.4184,0.803,0.4122); rgb(165pt)=(0.4322,0.8028,0.4013); rgb(166pt)=(0.4463,0.8024,0.3904); rgb(167pt)=(0.4608,0.8018,0.3797); rgb(168pt)=(0.4753,0.8011,0.3691); rgb(169pt)=(0.4899,0.8002,0.3586); rgb(170pt)=(0.5044,0.7993,0.348); rgb(171pt)=(0.5187,0.7982,0.3374); rgb(172pt)=(0.5329,0.797,0.3267); rgb(173pt)=(0.547,0.7957,0.3159); rgb(175pt)=(0.5748,0.7929,0.2941); rgb(176pt)=(0.5886,0.7913,0.2833); rgb(177pt)=(0.6024,0.7896,0.2726); rgb(178pt)=(0.6161,0.7878,0.2622); rgb(179pt)=(0.6297,0.7859,0.2521); rgb(180pt)=(0.6433,0.7839,0.2423); rgb(181pt)=(0.6567,0.7818,0.2329); rgb(182pt)=(0.6701,0.7796,0.2239); rgb(183pt)=(0.6833,0.7773,0.2155); rgb(184pt)=(0.6963,0.775,0.2075); rgb(185pt)=(0.7091,0.7727,0.1998); rgb(186pt)=(0.7218,0.7703,0.1924); rgb(187pt)=(0.7344,0.7679,0.1852); rgb(188pt)=(0.7468,0.7654,0.1782); rgb(189pt)=(0.759,0.7629,0.1717); rgb(190pt)=(0.771,0.7604,0.1658); rgb(191pt)=(0.7829,0.7579,0.1608); rgb(192pt)=(0.7945,0.7554,0.157); rgb(193pt)=(0.806,0.7529,0.1546); rgb(194pt)=(0.8172,0.7505,0.1535); rgb(195pt)=(0.8281,0.7481,0.1536); rgb(196pt)=(0.8389,0.7457,0.1546); rgb(197pt)=(0.8495,0.7435,0.1564); rgb(198pt)=(0.86,0.7413,0.1587); rgb(199pt)=(0.8703,0.7392,0.1615); rgb(200pt)=(0.8804,0.7372,0.165); rgb(201pt)=(0.8903,0.7353,0.1695); rgb(202pt)=(0.9,0.7336,0.1749); rgb(203pt)=(0.9093,0.7321,0.1815); rgb(204pt)=(0.9184,0.7308,0.189); rgb(205pt)=(0.9272,0.7298,0.1973); rgb(206pt)=(0.9357,0.729,0.2061); rgb(207pt)=(0.944,0.7285,0.2151); rgb(208pt)=(0.9523,0.7284,0.2237); rgb(209pt)=(0.9606,0.7285,0.2312); rgb(210pt)=(0.9689,0.7292,0.2373); rgb(211pt)=(0.977,0.7304,0.2418); rgb(212pt)=(0.9842,0.733,0.2446); rgb(213pt)=(0.99,0.7365,0.2429); rgb(214pt)=(0.9946,0.7407,0.2394); rgb(215pt)=(0.9966,0.7458,0.2351); rgb(216pt)=(0.9971,0.7513,0.2309); rgb(217pt)=(0.9972,0.7569,0.2267); rgb(218pt)=(0.9971,0.7626,0.2224); rgb(219pt)=(0.9969,0.7683,0.2181); rgb(220pt)=(0.9966,0.774,0.2138); rgb(221pt)=(0.9962,0.7798,0.2095); rgb(222pt)=(0.9957,0.7856,0.2053); rgb(223pt)=(0.9949,0.7915,0.2012); rgb(224pt)=(0.9938,0.7974,0.1974); rgb(225pt)=(0.9923,0.8034,0.1939); rgb(226pt)=(0.9906,0.8095,0.1906); rgb(227pt)=(0.9885,0.8156,0.1875); rgb(228pt)=(0.9861,0.8218,0.1846); rgb(229pt)=(0.9835,0.828,0.1817); rgb(230pt)=(0.9807,0.8342,0.1787); rgb(231pt)=(0.9778,0.8404,0.1757); rgb(232pt)=(0.9748,0.8467,0.1726); rgb(233pt)=(0.972,0.8529,0.1695); rgb(234pt)=(0.9694,0.8591,0.1665); rgb(235pt)=(0.9671,0.8654,0.1636); rgb(236pt)=(0.9651,0.8716,0.1608); rgb(237pt)=(0.9634,0.8778,0.1582); rgb(238pt)=(0.9619,0.884,0.1557); rgb(239pt)=(0.9608,0.8902,0.1532); rgb(240pt)=(0.9601,0.8963,0.1507); rgb(241pt)=(0.9596,0.9023,0.148); rgb(242pt)=(0.9595,0.9084,0.145); rgb(243pt)=(0.9597,0.9143,0.1418); rgb(244pt)=(0.9601,0.9203,0.1382); rgb(245pt)=(0.9608,0.9262,0.1344); rgb(246pt)=(0.9618,0.932,0.1304); rgb(247pt)=(0.9629,0.9379,0.1261); rgb(248pt)=(0.9642,0.9437,0.1216); rgb(249pt)=(0.9657,0.9494,0.1168); rgb(250pt)=(0.9674,0.9552,0.1116); rgb(251pt)=(0.9692,0.9609,0.1061); rgb(252pt)=(0.9711,0.9667,0.1001); rgb(253pt)=(0.973,0.9724,0.0938); rgb(254pt)=(0.9749,0.9782,0.0872); rgb(255pt)=(0.9769,0.9839,0.0805)},
xmin=-14,
xmax=2,
xlabel style={font=\bfseries\color{white!15!black}},
xlabel={NMSE in dB},
ymin=0,
ymax=800,
ytick={0,100,200,300,400,500,600,700,800},
ylabel style={font=\bfseries\color{white!15!black}},
ylabel={Number of network realizations},
axis background/.style={fill=white},
xmajorgrids,
ymajorgrids
]
\node[below right, align=left, font=\bfseries]
at (rel axis cs:0.6,0.95) {$\mu = 0.01$, $\ell = 50$, $\eta=50$\\Prediction horizon: 5 slots};
\addplot[area legend, table/row sep=crcr, patch, patch type=rectangle, shader=flat corner, draw=white!15!black, forget plot, patch table with point meta={%
1	2	3	4	1\\
6	7	8	9	1\\
11	12	13	14	1\\
16	17	18	19	1\\
21	22	23	24	1\\
26	27	28	29	1\\
31	32	33	34	1\\
36	37	38	39	1\\
41	42	43	44	1\\
46	47	48	49	1\\
51	52	53	54	1\\
56	57	58	59	1\\
61	62	63	64	1\\
66	67	68	69	1\\
71	72	73	74	1\\
76	77	78	79	1\\
81	82	83	84	1\\
86	87	88	89	1\\
91	92	93	94	1\\
96	97	98	99	1\\
101	102	103	104	1\\
106	107	108	109	1\\
111	112	113	114	1\\
116	117	118	119	1\\
121	122	123	124	1\\
126	127	128	129	1\\
131	132	133	134	1\\
136	137	138	139	1\\
141	142	143	144	1\\
146	147	148	149	1\\
151	152	153	154	1\\
156	157	158	159	1\\
161	162	163	164	1\\
166	167	168	169	1\\
171	172	173	174	1\\
176	177	178	179	1\\
181	182	183	184	1\\
186	187	188	189	1\\
191	192	193	194	1\\
196	197	198	199	1\\
201	202	203	204	1\\
206	207	208	209	1\\
211	212	213	214	1\\
216	217	218	219	1\\
221	222	223	224	1\\
226	227	228	229	1\\
231	232	233	234	1\\
236	237	238	239	1\\
241	242	243	244	1\\
246	247	248	249	1\\
}]
table[row sep=crcr] {%
x	y\\
-13.7930248949699	0\\
-13.7930248949699	0\\
-13.7930248949699	2\\
-13.4768225093788	2\\
-13.4768225093788	0\\
-13.4768225093788	0\\
-13.4768225093788	0\\
-13.4768225093788	5\\
-13.1606201237876	5\\
-13.1606201237876	0\\
-13.1606201237876	0\\
-13.1606201237876	0\\
-13.1606201237876	13\\
-12.8444177381965	13\\
-12.8444177381965	0\\
-12.8444177381965	0\\
-12.8444177381965	0\\
-12.8444177381965	15\\
-12.5282153526053	15\\
-12.5282153526053	0\\
-12.5282153526053	0\\
-12.5282153526053	0\\
-12.5282153526053	22\\
-12.2120129670142	22\\
-12.2120129670142	0\\
-12.2120129670142	0\\
-12.2120129670142	0\\
-12.2120129670142	50\\
-11.895810581423	50\\
-11.895810581423	0\\
-11.895810581423	0\\
-11.895810581423	0\\
-11.895810581423	87\\
-11.5796081958318	87\\
-11.5796081958318	0\\
-11.5796081958318	0\\
-11.5796081958318	0\\
-11.5796081958318	128\\
-11.2634058102407	128\\
-11.2634058102407	0\\
-11.2634058102407	0\\
-11.2634058102407	0\\
-11.2634058102407	180\\
-10.9472034246495	180\\
-10.9472034246495	0\\
-10.9472034246495	0\\
-10.9472034246495	0\\
-10.9472034246495	252\\
-10.6310010390584	252\\
-10.6310010390584	0\\
-10.6310010390584	0\\
-10.6310010390584	0\\
-10.6310010390584	329\\
-10.3147986534672	329\\
-10.3147986534672	0\\
-10.3147986534672	0\\
-10.3147986534672	0\\
-10.3147986534672	434\\
-9.99859626787604	434\\
-9.99859626787604	0\\
-9.99859626787604	0\\
-9.99859626787604	0\\
-9.99859626787604	517\\
-9.68239388228488	517\\
-9.68239388228488	0\\
-9.68239388228488	0\\
-9.68239388228488	0\\
-9.68239388228488	646\\
-9.36619149669372	646\\
-9.36619149669372	0\\
-9.36619149669372	0\\
-9.36619149669372	0\\
-9.36619149669372	649\\
-9.04998911110256	649\\
-9.04998911110256	0\\
-9.04998911110256	0\\
-9.04998911110256	0\\
-9.04998911110256	760\\
-8.7337867255114	760\\
-8.7337867255114	0\\
-8.7337867255114	0\\
-8.7337867255114	0\\
-8.7337867255114	715\\
-8.41758433992024	715\\
-8.41758433992024	0\\
-8.41758433992024	0\\
-8.41758433992024	0\\
-8.41758433992024	722\\
-8.10138195432908	722\\
-8.10138195432908	0\\
-8.10138195432908	0\\
-8.10138195432908	0\\
-8.10138195432908	676\\
-7.78517956873792	676\\
-7.78517956873792	0\\
-7.78517956873792	0\\
-7.78517956873792	0\\
-7.78517956873792	618\\
-7.46897718314676	618\\
-7.46897718314676	0\\
-7.46897718314676	0\\
-7.46897718314676	0\\
-7.46897718314676	623\\
-7.1527747975556	623\\
-7.1527747975556	0\\
-7.1527747975556	0\\
-7.1527747975556	0\\
-7.1527747975556	526\\
-6.83657241196445	526\\
-6.83657241196445	0\\
-6.83657241196445	0\\
-6.83657241196445	0\\
-6.83657241196445	455\\
-6.52037002637329	455\\
-6.52037002637329	0\\
-6.52037002637329	0\\
-6.52037002637329	0\\
-6.52037002637329	352\\
-6.20416764078213	352\\
-6.20416764078213	0\\
-6.20416764078213	0\\
-6.20416764078213	0\\
-6.20416764078213	305\\
-5.88796525519097	305\\
-5.88796525519097	0\\
-5.88796525519097	0\\
-5.88796525519097	0\\
-5.88796525519097	240\\
-5.57176286959981	240\\
-5.57176286959981	0\\
-5.57176286959981	0\\
-5.57176286959981	0\\
-5.57176286959981	175\\
-5.25556048400865	175\\
-5.25556048400865	0\\
-5.25556048400865	0\\
-5.25556048400865	0\\
-5.25556048400865	121\\
-4.93935809841749	121\\
-4.93935809841749	0\\
-4.93935809841749	0\\
-4.93935809841749	0\\
-4.93935809841749	126\\
-4.62315571282633	126\\
-4.62315571282633	0\\
-4.62315571282633	0\\
-4.62315571282633	0\\
-4.62315571282633	70\\
-4.30695332723517	70\\
-4.30695332723517	0\\
-4.30695332723517	0\\
-4.30695332723517	0\\
-4.30695332723517	41\\
-3.99075094164401	41\\
-3.99075094164401	0\\
-3.99075094164401	0\\
-3.99075094164401	0\\
-3.99075094164401	34\\
-3.67454855605285	34\\
-3.67454855605285	0\\
-3.67454855605285	0\\
-3.67454855605285	0\\
-3.67454855605285	30\\
-3.35834617046169	30\\
-3.35834617046169	0\\
-3.35834617046169	0\\
-3.35834617046169	0\\
-3.35834617046169	21\\
-3.04214378487053	21\\
-3.04214378487053	0\\
-3.04214378487053	0\\
-3.04214378487053	0\\
-3.04214378487053	17\\
-2.72594139927938	17\\
-2.72594139927938	0\\
-2.72594139927937	0\\
-2.72594139927937	0\\
-2.72594139927937	7\\
-2.40973901368822	7\\
-2.40973901368822	0\\
-2.40973901368822	0\\
-2.40973901368822	0\\
-2.40973901368822	6\\
-2.09353662809706	6\\
-2.09353662809706	0\\
-2.09353662809706	0\\
-2.09353662809706	0\\
-2.09353662809706	4\\
-1.7773342425059	4\\
-1.7773342425059	0\\
-1.7773342425059	0\\
-1.7773342425059	0\\
-1.7773342425059	8\\
-1.46113185691474	8\\
-1.46113185691474	0\\
-1.46113185691474	0\\
-1.46113185691474	0\\
-1.46113185691474	3\\
-1.14492947132358	3\\
-1.14492947132358	0\\
-1.14492947132358	0\\
-1.14492947132358	0\\
-1.14492947132358	3\\
-0.828727085732419	3\\
-0.828727085732419	0\\
-0.828727085732419	0\\
-0.828727085732419	0\\
-0.828727085732419	0\\
-0.51252470014126	0\\
-0.51252470014126	0\\
-0.512524700141259	0\\
-0.512524700141259	0\\
-0.512524700141259	4\\
-0.1963223145501	4\\
-0.1963223145501	0\\
-0.1963223145501	0\\
-0.1963223145501	0\\
-0.1963223145501	0\\
0.119880071041059	0\\
0.119880071041059	0\\
0.119880071041058	0\\
0.119880071041058	0\\
0.119880071041058	1\\
0.436082456632217	1\\
0.436082456632217	0\\
0.436082456632217	0\\
0.436082456632217	0\\
0.436082456632217	1\\
0.752284842223377	1\\
0.752284842223377	0\\
0.752284842223377	0\\
0.752284842223377	0\\
0.752284842223377	1\\
1.06848722781454	1\\
1.06848722781454	0\\
1.06848722781454	0\\
1.06848722781454	0\\
1.06848722781454	2\\
1.3846896134057	2\\
1.3846896134057	0\\
1.38468961340569	0\\
1.38468961340569	0\\
1.38468961340569	0\\
1.70089199899685	0\\
1.70089199899685	0\\
1.70089199899685	0\\
1.70089199899685	0\\
1.70089199899685	4\\
2.01709438458801	4\\
2.01709438458801	0\\
2.01709438458801	0\\
};
\end{axis}

\end{tikzpicture}%
 }
\caption{${\rm NMSE}$ distribution across network realizations. Results correspond to the case labeled $\ell=50$ in Fig.~\ref{Fig:PredictorMsgs} with a prediction horizon of five time~slots.} 
\label{Fig:PredError}
\end{figure}

Recalling that the model used for prediction is independent of the individual realizations of the interference process, Fig.~\ref{Fig:PredError} explores how the prediction error deviates from its average value in Figs.~\ref{Fig:Predictor2} and \ref{Fig:PredictorMsgs} for different realizations of the network. Despite the fact that significant deviations exist for some realizations, most concentrate in the vicinity of the mean error.

\subsection{Sensitivity to Other System Models}

Now that we have seen that the interference predictor works well in the base system  used for its design, it is of interest whether it also works in systems with other models. In other words, we are interested in how sensitive or robust the predictor is to variations or errors in the assumptions. 

\smallskip
\subsubsection{System Model}
We evaluate the predictor performance in \eqref{eq:Kalman_Predict} for the following models:  The node locations are no longer sampled from a homogeneous PPP but show some inhomogeneity. The experienced small-scale fading is no longer Rayleigh but is rather Nakagami distributed.  The message lengths are no longer fixed to $\msglen$ but are sampled from a Poisson distribution with \mbox{parameter $\msglen$.}  Nodes no longer transmit independently of the channel conditions but rather depend on the channel activity.

\begin{figure}[htbp]
 \centering
  \includegraphics[width=.49\textwidth,keepaspectratio=true]{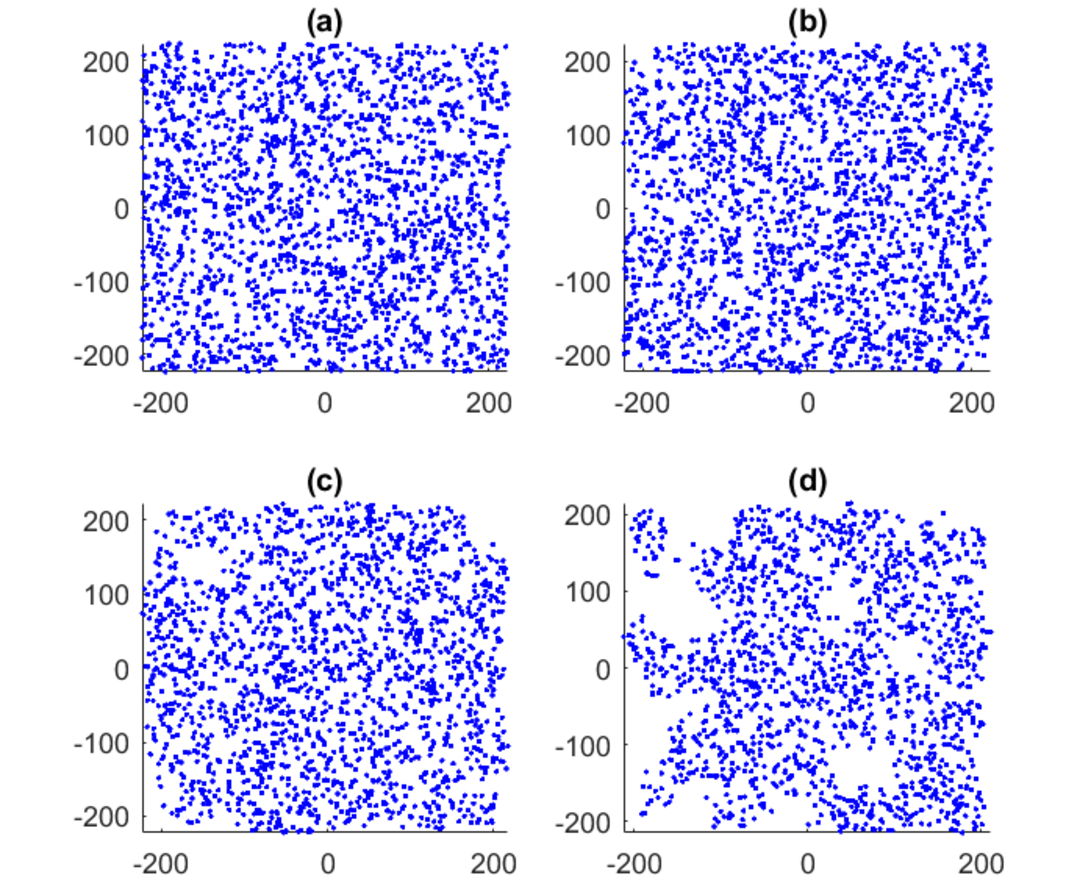}
\caption{Realizations of clustered node locations over an area of 200,000 square units. The original PPP (a) has an intensity $\lambda=0.01$. The clustered nodes are generated by thinning this realization with distance parameter $r=40$ and neighbor cardinality of (b) $k=20$, (c) $k=30$, and (d) $k=40$.} 
\label{Fig:Clustering}
\end{figure}  

The modeling of the locations is generalized by using an inhomogeneous random node distribution, in which nodes can be clustered in groups or concentrate at hotspots. This is achieved by thinning the original PPP realizations, such that nodes with at least a certain number $k$ of neighboring nodes within a distance $r$ are kept~\cite{bettstetter07:mswim}. Some examples are shown in~Fig.~\ref{Fig:Clustering} with fixed $r$ and increasing $k$. Changes to the node density resulting from this thinning do neither affect the interference correlation nor the predictor design, as \eqref{eq:AutocorrFunction} does not contain~$\lambda$. The small-scale fading is generalized using the Nakagami distribution. Rayleigh fading, which was used for parameterization, corresponds to Nakagami fading with parameter $m=1$, and we evaluate two larger $m$-values representing environments with fewer reflections. We apply a block fading channel model, where the channel states remain constant during the channel coherence time, and change to independent values afterward. Many wireless technologies implement channel sensing to minimize packet collisions \cite{Rev3_a,Rev3_b}. Therefore, it is relevant to investigate the impact of transmissions allowed only if the sensed channel activity is below a given threshold from the transmission power. 

\begin{figure}[t]\centering
	\resizebox{.49\textwidth}{!}{
   \definecolor{mycolor1}{rgb}{0.39216,0.83137,0.07451}%
\begin{tikzpicture}

\begin{axis}[%
width=3.442in,
height=1.267in,
at={(0.577in,5.308in)},
scale only axis,
xmin=1,
xmax=10,
xtick={ 1,  2,  3,  4,  5,  6,  7,  8,  9, 10},
ymin=-20,
ymax=0,
axis background/.style={fill=white},
title style={font=\bfseries\small},
title={(a) Node distribution},
axis x line*=bottom,
axis y line*=left,
xmajorgrids,
ymajorgrids,
legend style={font=\footnotesize,at={(0.547,0)}, anchor=south west, legend cell align=left, align=left, draw=white!12!black}
]
\addplot [color=blue, line width=1.0pt]
  table[row sep=crcr]{%
1	-6.79889043075663\\
2	-5.25595632112754\\
3	-3.58983450918134\\
4	-2.91801545118356\\
5	-2.07284381352991\\
6	-1.74101034460547\\
7	-1.38086928052759\\
8	-1.1065258431944\\
9	-0.8358420245209\\
10	-0.824975345059205\\
};
\addlegendentry{No thinning}

\addplot [color=red, dashed, line width=1.0pt]
  table[row sep=crcr]{%
1	-6.76647732777952\\
2	-5.2175258141589\\
3	-3.53285315310874\\
4	-2.88690241724986\\
5	-2.0367268843024\\
6	-1.69592488285463\\
7	-1.32311501792427\\
8	-1.06259092217419\\
9	-0.784771336451653\\
10	-0.793247901841113\\
};
\addlegendentry{r=40 / k=30 thinning}

\addplot [color=mycolor1, dashdotted, line width=1.0pt]
  table[row sep=crcr]{%
1	-6.62754030939685\\
2	-5.07568012527989\\
3	-3.40123288152322\\
4	-2.73505201628435\\
5	-1.85736858900121\\
6	-1.53349979615188\\
7	-1.16774342314979\\
8	-0.91325436689853\\
9	-0.645879273945592\\
10	-0.632824580064577\\
};
\addlegendentry{r=40 / k=40 thinning}

\addplot [color=blue, line width=1.0pt, forget plot]
  table[row sep=crcr]{%
1	-17.5973222133454\\
2	-15.6549462022603\\
3	-14.1174486914474\\
4	-12.8770051920229\\
5	-11.8899139553537\\
6	-11.0330385216149\\
7	-10.3282115556114\\
8	-9.71871562410503\\
9	-9.10454805283693\\
10	-8.57954550748716\\
};
\addplot [color=red, dashed, line width=1.0pt, forget plot]
  table[row sep=crcr]{%
1	-17.6928127348448\\
2	-15.4327331198588\\
3	-13.9324902875487\\
4	-12.8114741233968\\
5	-11.812983746711\\
6	-10.931833787876\\
7	-10.2571064007353\\
8	-9.57642602247103\\
9	-8.96353404684547\\
10	-8.49343315286199\\
};
\addplot [color=mycolor1, dashdotted, line width=1.0pt, forget plot]
  table[row sep=crcr]{%
1	-17.3170408771678\\
2	-15.1426746514575\\
3	-13.6499506171062\\
4	-12.4556439559294\\
5	-11.4832265831028\\
6	-10.5577827931466\\
7	-9.86285135507433\\
8	-9.27920930530549\\
9	-8.6634542900841\\
10	-8.13599845114332\\
};
\end{axis}

\begin{axis}[%
width=3.442in,
height=1.267in,
at={(0.577in,3.538in)},
scale only axis,
xmin=1,
xmax=10,
xtick={ 1,  2,  3,  4,  5,  6,  7,  8,  9, 10},
ymin=-20,
ymax=0,
axis background/.style={fill=white},
title style={font=\bfseries\small},
title={(b) Fading model},
axis x line*=bottom,
axis y line*=left,
xmajorgrids,
ymajorgrids,
legend style={font=\footnotesize,at={(0.65,0)}, anchor=south west, legend cell align=left, align=left, draw=white!12!black}
]
\addplot [color=blue, line width=1.0pt]
  table[row sep=crcr]{%
1	-6.80363990203899\\
2	-5.29015268642964\\
3	-3.58640141267054\\
4	-2.93332838277497\\
5	-2.07339363451763\\
6	-1.75942890166806\\
7	-1.40877677716279\\
8	-1.1117758385442\\
9	-0.869348469374772\\
10	-0.828430790862431\\
};
\addlegendentry{Rayleigh}

\addplot [color=red, dashed, line width=1.0pt]
  table[row sep=crcr]{%
1	-6.73319132646131\\
2	-5.24725458407365\\
3	-3.62950543781672\\
4	-3.00457505636767\\
5	-2.18191647248648\\
6	-1.86131960126725\\
7	-1.53357297457505\\
8	-1.26571424321564\\
9	-1.04139434030747\\
10	-1.04082694935264\\
};
\addlegendentry{Nakagami m=4}

\addplot [color=mycolor1, dashdotted, line width=1.0pt]
  table[row sep=crcr]{%
1	-6.92338183570516\\
2	-5.46068558667754\\
3	-3.82222540977025\\
4	-3.17937758027529\\
5	-2.37357601981086\\
6	-2.04891243131892\\
7	-1.70488993041387\\
8	-1.41711574722805\\
9	-1.15931029753234\\
10	-1.19003543362047\\
};
\addlegendentry{Nakagami m=4}

\addplot [color=blue, line width=1.0pt, forget plot]
  table[row sep=crcr]{%
1	-17.6753912830455\\
2	-15.5660133176578\\
3	-14.1010878847707\\
4	-12.9122012271837\\
5	-11.9448850402068\\
6	-11.1265216847082\\
7	-10.3180088347838\\
8	-9.64909687250052\\
9	-9.12464961520759\\
10	-8.57366349294535\\
};
\addplot [color=red, dashed, line width=1.0pt, forget plot]
  table[row sep=crcr]{%
1	-16.6139298670954\\
2	-14.4394995427185\\
3	-13.0829471051857\\
4	-12.0069211912218\\
5	-11.1929366684847\\
6	-10.4827113218886\\
7	-9.94861621668077\\
8	-9.38913998691154\\
9	-8.95768066722369\\
10	-8.56565631843565\\
};
\addplot [color=mycolor1, dashdotted, line width=1.0pt, forget plot]
  table[row sep=crcr]{%
1	-17.4233862531871\\
2	-15.3905392908505\\
3	-14.0120062703625\\
4	-12.9692638024666\\
5	-12.184667955936\\
6	-11.4927597193911\\
7	-10.8557919898058\\
8	-10.3703749380483\\
9	-9.91250463581659\\
10	-9.53286954976984\\
};
\end{axis}

\begin{axis}[%
width=3.442in,
height=1.267in,
at={(0.577in,1.769in)},
scale only axis,
xmin=1,
xmax=10,
xtick={ 1,  2,  3,  4,  5,  6,  7,  8,  9, 10},
ymin=-20,
ymax=0,
axis background/.style={fill=white},
title style={font=\bfseries\small},
title={(c) Traffic model},
axis x line*=bottom,
axis y line*=left,
xmajorgrids,
ymajorgrids,
legend style={font=\footnotesize,at={(0.65,0)}, anchor=south west, legend cell align=left, align=left, draw=white!12!black}
]
\addplot [color=blue, line width=1.0pt]
  table[row sep=crcr]{%
1	-6.80363990203899\\
2	-5.29015268642964\\
3	-3.58640141267054\\
4	-2.93332838277497\\
5	-2.07339363451763\\
6	-1.75942890166806\\
7	-1.40877677716279\\
8	-1.1117758385442\\
9	-0.869348469374772\\
10	-0.828430790862431\\
};
\addlegendentry{Fixed length}

\addplot [color=red, dashed, line width=1.0pt]
  table[row sep=crcr]{%
1	-6.96033444031506\\
2	-4.99278593538829\\
3	-3.54867322814539\\
4	-2.88250619532518\\
5	-2.0935262387025\\
6	-1.72509710157866\\
7	-1.3975962833856\\
8	-1.1231483269392\\
9	-0.948072106827695\\
10	-0.949482147285496\\
};
\addlegendentry{Poisson length}

\addplot [color=blue, line width=1.0pt, forget plot]
  table[row sep=crcr]{%
1	-17.6753912830455\\
2	-15.5660133176578\\
3	-14.1010878847707\\
4	-12.9122012271837\\
5	-11.9448850402068\\
6	-11.1265216847082\\
7	-10.3180088347838\\
8	-9.64909687250052\\
9	-9.12464961520759\\
10	-8.57366349294535\\
};
\addplot [color=red, dashed, line width=1.0pt, forget plot]
  table[row sep=crcr]{%
1	-17.704922973817\\
2	-15.579458443684\\
3	-14.1420350909904\\
4	-12.9054901248249\\
5	-11.9443976711532\\
6	-11.1827414703377\\
7	-10.380042720919\\
8	-9.69017964388409\\
9	-9.06134675751933\\
10	-8.62270361419282\\
};
\end{axis}

\begin{axis}[%
width=3.442in,
height=1.267in,
at={(0.577in,0.0in)},
scale only axis,
xmin=1,
xmax=10,
xtick={ 1,  2,  3,  4,  5,  6,  7,  8,  9, 10},
xlabel style={font=\bfseries\color{white!15!black}},
xlabel={Prediction horizon in slots},
ymin=-20,
ymax=0,
axis background/.style={fill=white},
title style={font=\bfseries\small},
title={(d) Channel sensing},
axis x line*=bottom,
axis y line*=left,
xmajorgrids,
ymajorgrids,
legend style={font=\footnotesize,at={(0.5,0)}, anchor=south west, legend cell align=left, align=left, draw=white!12!black}
]
\addplot [color=blue, line width=1.0pt]
  table[row sep=crcr]{%
1	-6.80363990203899\\
2	-5.29015268642964\\
3	-3.58640141267054\\
4	-2.93332838277497\\
5	-2.07339363451763\\
6	-1.75942890166806\\
7	-1.40877677716279\\
8	-1.1117758385442\\
9	-0.869348469374772\\
10	-0.828430790862431\\
};
\addlegendentry{No channel sensing}

\addplot [color=red, dashed, line width=1.0pt]
  table[row sep=crcr]{%
1	-17.2998676306643\\
2	-14.9586792211837\\
3	-13.3900293497177\\
4	-12.1489831537844\\
5	-11.1747397143544\\
6	-10.3718207804481\\
7	-9.69608646019185\\
8	-9.03239644485987\\
9	-8.47212485125958\\
10	-7.9706377209443\\
};
\addlegendentry{Sensing threshold -20 dB}

\addplot [color=mycolor1, dashdotted, line width=1.0pt]
  table[row sep=crcr]{%
1	-15.8113492243422\\
2	-13.6722661840501\\
3	-12.2912155760187\\
4	-11.1832797103919\\
5	-10.2531990230456\\
6	-9.40851739206306\\
7	-8.7800497367091\\
8	-8.1936185174429\\
9	-7.6324072501147\\
10	-7.16670972987385\\
};
\addlegendentry{Sensing threshold -50 dB}

\addplot [color=red, dashed, line width=1.0pt, forget plot]
  table[row sep=crcr]{%
1	-5.12486673703599\\
2	-4.11684297195253\\
3	-2.67640046933791\\
4	-2.17373080799935\\
5	-1.37828433050487\\
6	-1.1189395414944\\
7	-0.82553930010517\\
8	-0.600798301396932\\
9	-0.540225965252203\\
10	-0.520176688519468\\
};
\addplot [color=mycolor1, dashdotted, line width=1.0pt, forget plot]
  table[row sep=crcr]{%
1	-4.89558732668427\\
2	-3.9753125287206\\
3	-2.57018994373626\\
4	-2.08171908004534\\
5	-1.29421659827012\\
6	-1.04606292555149\\
7	-0.72710773904959\\
8	-0.541833905377642\\
9	-0.482852483338554\\
10	-0.483348391504563\\
};
\addplot [color=blue, line width=1.0pt, forget plot]
  table[row sep=crcr]{%
1	-17.5973222133454\\
2	-15.6549462022603\\
3	-14.1174486914474\\
4	-12.8770051920229\\
5	-11.8899139553537\\
6	-11.0330385216149\\
7	-10.3282115556114\\
8	-9.71871562410503\\
9	-9.10454805283693\\
10	-8.57954550748716\\
};
\end{axis}

\begin{axis}[%
width=4in,
height=7in,
at={(0.3in,0in)},
scale only axis,
xmin=0,
xmax=1,
ymin=0,
ymax=1,
ylabel style={font=\bfseries\color{white!15!black}},
ylabel={NMSE in dB},
axis line style={draw=none},
ticks=none,
axis x line*=bottom,
axis y line*=left
]
\node[below right, align=left, font=\bfseries]
at (rel axis cs:0.1,0.93) {$\ell = 10$};
\node[below right, align=left, font=\bfseries]
at (rel axis cs:0.75,0.9) {$\ell = 100$};
\node[below right, align=left, font=\bfseries]
at (rel axis cs:0.1,0.67) {$\ell = 10$};
\node[below right, align=left, font=\bfseries]
at (rel axis cs:0.7,0.64) {$\ell = 100$};
\node[below right, align=left, font=\bfseries]
at (rel axis cs:0.69,0.34) {$\ell = 100$};
\node[below right, align=left, font=\bfseries]
at (rel axis cs:0.33,0.40) {$\ell = 10$};
\node[below right, align=left, font=\bfseries]
at (rel axis cs:0.25,0.15) {$\ell = 10$};
\node[below right, align=left, font=\bfseries]
at (rel axis cs:0.75,0.14) {$\ell = 100$};
\end{axis}
\end{tikzpicture}%
 }
\caption{Sensitivity of the predictor against (a) inhomogeneous node distribution (b) non-Rayleigh fading, (c) Poisson distributed message lengths, and (d) transmissions depending on channel activity.
Results labeled $\ell=10$ correspond to Setup~1 from Table~\ref{tab:Scenarios}. Those labeled $\ell=100$ extend the message length of that setup. The interference coherence time for $\theta = 0.25$ is $5$ and $47$ time slots for $\ell = 10$ and $100$, respectively.
} 
\label{Fig:Sensitivity_1}
\end{figure}
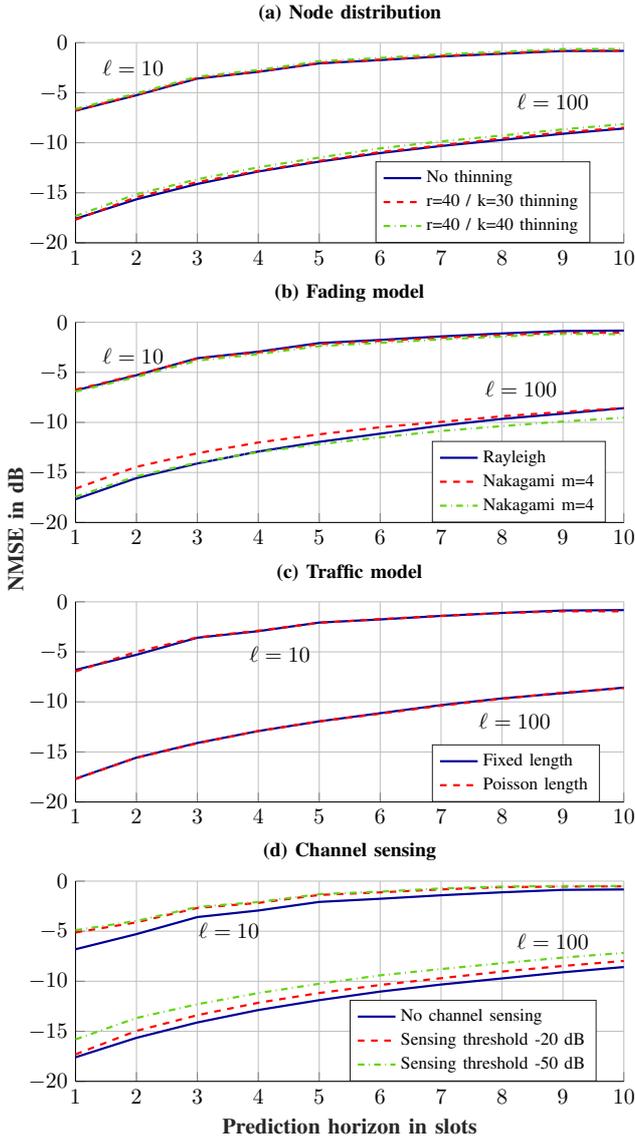 

\smallskip
\subsubsection{Performance Results}
Figure \ref{Fig:Sensitivity_1} shows the NMSE of the interference predictor for these setups if the interference samples at its input follow these alternative node placement, fading, and traffic models. Subfigure~(a) shows the impact of the design mismatch in terms of the inhomogeneous node distribution. The predictor exhibits a robust behavior for significantly high in\-homogeneity.  Deviations are marginal and can only be appreciated for $k= 40$ (Fig.~\ref{Fig:Clustering}~(d)). 
Subfigure~(b) assesses the impact of a mismatch in the fading model, which the evaluation shows to be more significant for larger messages, although still close to the perfect match (i.e., Rayleigh). Subfigure~(c) reveals that the predictor is robust against the traffic model as long as the mean message length remains unchanged. The deviations are within the averaging noise. Nevertheless, Subfigure~(d) indicates that the predictor is sensitive to mismatches in the channel access mechanism. Performance degrades as more intended transmissions are prevented to start (i.e., lower thresholds). The interference model should incorporate these mechanisms for best performance. 

\subsection{Evaluation for Specific Technologies}

We now study the prediction performance for parameter values that are typical in two wireless technologies: Long Term Evolution (LTE) used for cellular systems \cite{LTEStandard} and IEEE~802.15.4 used for wireless sensor networks (WSN)\cite{WirelesSensorStandard}. This evaluation will highlight the versatility of the predictor. A message length of 20~slots is used in all scenarios.
\smallskip

\subsubsection{System Parameters}
We consider an LTE downlink with carrier frequency~$f_c$ in the  $2$~GHz band. A base station assigns the mobile nodes to a specific sub\-carrier (or block of contiguous sub\-carriers). Scheduling decisions can be made every millisecond, defining a slot duration of one millisecond~\cite{LTEStandard}. Each transmission experiences interference from the base stations of neighboring cells that are using the same frequency band (inter-cell interference). No exclusion area is modelled and hence interferers can be located arbitrarily close to the receiver. Although interferers are not mobile in this scenario, interference is impacted by node mobility leading to similar effects.
In order to harness predictions at the base station, a prediction horizon above five slots is required to compensate for the processing delays involved~\cite{schmidt2011bit}. We model the mobility with a maximum speed $\nu_{\rm max} = 1.67~{\rm m/s}=6~{\rm km/h}$ for pedestrians and $40$ and $80~{\rm km/h}$ for vehicles. 
We also consider a WSN with $f_c=2.4$~GHz. The standard defines message lengths with a minimum of eight packets. With a packet duration of $577$~$\mu$s, the slot duration is $4.6$~ms~\cite{WirelesSensorStandard}. These values give a prediction horizon of $23$~ms for predicting five slots ahead. We use the same pedestrian mobility as in LTE. All parameters are summarized in Table~\ref{tab:Technologies}. The maximum Doppler shift is $\Delta_f = 2 f_c\, \nu_{\rm max}/c$ with the speed of light $c$. 

\begin{table}[tp]
\renewcommand{\arraystretch}{1.3}
\begin{center}
\caption{System Parameters for LTE and WSN Scenarios}
\label{tab:Technologies}
\begin{tabular}{ccccc}
\hline
{Scenario} & Speed  & Traffic pa- &  Channel coher- & Doppler \\
 & $\nu$ & rameter $\txp$ & ence time $\chlen$ &  shift $\Delta_f^*$\\\hline
{LTE 1} & ~6 km/h & 0.01 / slot & 225 slots & 0.022 \\ 
{LTE 2} & 40 km/h & 0.01 / slot & ~35 slots & 0.148 \\ 
{LTE 3} & 80 km/h & 0.01 / slot & ~17 slots & 0.296 \\ 
{WSN}   & ~6 km/h & 0.01 / slot & ~50 slots & 0.125 \\ \hline
\multicolumn{5}{r}{The $\Delta_f$ is normalized to the system slot frequency.} \\ 
\end{tabular}
\end{center}
\end{table}

\smallskip
\subsubsection{Performance Results}
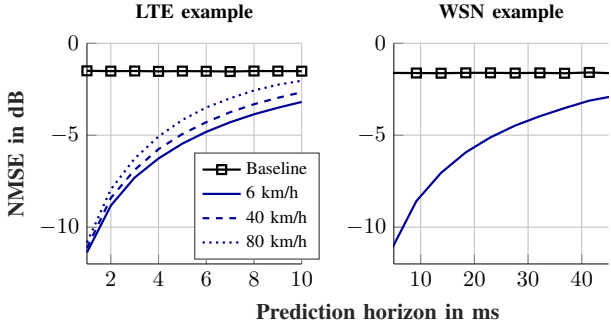
\begin{figure}[htbp]
 \centering
  \resizebox{.49\textwidth}{!}{
   \begin{tikzpicture}

\begin{axis}[%
width=1.952in,
height=2.0in,
at={(0.1in,0.2in)},
xmin=1,
xmax=10,
ymin=-12,
ymax=0,
ylabel style={font=\bfseries\color{white!15!black}},
ylabel={NMSE in dB},
axis background/.style={fill=white},
title style={font=\bfseries\small},
title={LTE example},
axis x line*=bottom,
axis y line*=left,
xmajorgrids,
ymajorgrids,
legend style={font=\footnotesize,at={(0.5,0.02)}, anchor=south west, legend cell align=left, align=left, draw=white!15!black}
]

\addplot [color=black, line width=1.0pt, mark=square, mark options={solid, black}]
  table[row sep=crcr]{%
1	-1.49953520872856\\
2	-1.5159839333087\\
3	-1.50921046749691\\
4	-1.52855190285019\\
5	-1.51356631015462\\
6	-1.52234408123055\\
7	-1.53796624412394\\
8	-1.51046669740013\\
9	-1.51344092071456\\
10	-1.51834507159688\\
};
\addlegendentry{Baseline}

\addplot [color=blue, line width=1.0pt]
  table[row sep=crcr]{%
1	-11.3762987250163\\
2	-8.82838421452315\\
3	-7.30259702216771\\
4	-6.26362057256036\\
5	-5.45729161540989\\
6	-4.81661570486572\\
7	-4.29685207461339\\
8	-3.86265856761719\\
9	-3.50660259825519\\
10	-3.19151205703097\\
};
\addlegendentry{6 km/h}

\addplot [color=blue, dashed, line width=1.0pt]
  table[row sep=crcr]{%
1	-11.1387343695366\\
2	-8.4151936900447\\
3	-6.88244463854983\\
4	-5.75760917685898\\
5	-4.9360983207588\\
6	-4.29073232584895\\
7	-3.75080098969588\\
8	-3.31597952643267\\
9	-2.96754758877559\\
10	-2.66148003631383\\
};
\addlegendentry{40 km/h}

\addplot [color=blue, dotted, line width=1.0pt]
  table[row sep=crcr]{%
1	-10.8357536854438\\
2	-7.94846539527997\\
3	-6.26691083889618\\
4	-5.06931452294685\\
5	-4.16851655893811\\
6	-3.49293530491292\\
7	-2.99271630160764\\
8	-2.56588008546348\\
9	-2.25577374116068\\
10	-2.03814870557784\\
};
\addlegendentry{80 km/h}
\end{axis}

\begin{axis}[%
width=1.952in,
height=2.0in,
at={(2in,0.2in)},
xmin=5,
xmax=45,
ymin=-12,
ymax=0,
axis background/.style={fill=white},
title style={font=\bfseries\small},
title={WSN example},
axis x line*=bottom,
axis y line*=left,
xmajorgrids,
ymajorgrids
]
\addplot [color=black, line width=1.0pt, mark=square, mark options={solid, black}]
  table[row sep=crcr]{%
4.6	-1.61370022367284\\
9.2	-1.61858509696653\\
13.8	-1.62942204775027\\
18.4	-1.60890526696798\\
23	-1.60908070030418\\
27.6	-1.62091863096039\\
32.2	-1.61069597171055\\
36.8	-1.63188884128819\\
41.4	-1.58854565872581\\
46	-1.63371154949399\\
};

\addplot [color=blue, line width=1.0pt]
  table[row sep=crcr]{%
4.6	-11.1928164953454\\
9.2	-8.58143312759029\\
13.8	-7.04714568649398\\
18.4	-5.93177168065552\\
23	-5.11711597669242\\
27.6	-4.47588486289094\\
32.2	-3.96774846959568\\
36.8	-3.52709823709046\\
41.4	-3.12237287250842\\
46	-2.86821770761205\\
};
\end{axis}

\begin{axis}[%
width=3.6in,
height=2.0in,
at={(0.1in,0in)},
scale only axis,
xmin=0,
xmax=1,
xlabel style={font=\bfseries\color{white!15!black}},
xlabel={Prediction horizon in ms},
ymin=0,
ymax=1,
axis line style={draw=none},
ticks=none,
axis x line*=bottom,
axis y line*=left
]
\end{axis}
\end{tikzpicture}%
 }
\caption{Predictor performance for the different LTE and WSN scenarios from Table~\ref{tab:Technologies}. The interference coherence time for $\theta = 0.25$ is $10$, $9$, and $7$ slots for speeds of $6$, $40$, and $80$ km/h for LTE, and $9$ slots for the WSN.} 
\label{Fig:Technology}
\end{figure}

Fig.~\ref{Fig:Technology} shows the ultimate predictor performance with these parameters if we map the prediction horizons to the signal timing and expected channel dynamics. We find that it is possible to predict beyond five slots with an error below the mean interference baseline in both cases. The gains against the channel and the last value predictor are in line with those above. 

\subsection{Evaluation for Coexisting Technologies}

A setup with multiple wireless technologies coexisting in the same band contributes to more efficient use of scarce resources (see \cite{Rev3_a,Rev3_b}). In this context, we explore how the proposed predictor is affected by interference from coexisting technologies not accounted for in its design. Specifically, we consider the coexistence of IEEE~802.11ax \mbox{Wi-Fi} and LTE in the $5$~GHz band \cite{Rev3_b}. The predictor is designed for predicting the interference arising from the Wi-Fi deployment alone and its performance is affected by the interference generated by the LTE system  operating in the same band. 

\smallskip
\subsubsection{System Parameters}
Wi-Fi is operated in the single-user (SU) mode. Users and access points compete for a complete $20$~MHz channel, following a simplified clear channel assessment (CCA) protocol. Channel access opportunities are arranged in $0.5$~ms slots, at which users transmit with $\mu=0.01$. Before initiating a transmission, the channel is sensed. If the sensed power is at least $50$~dB (sensing threshold) below the transmit power, the channel is deemed idle and transmission begins. Otherwise, the transmission is discarded. The predictor design did not take such channel sensing into account. 

The coexisting LTE system is considered in two modes: unlicensed spectrum (LTE-U) and licensed assisted access (LTE-LAA). 
In \mbox{LTE-U}, the base stations  transmit without channel sensing. To mitigate their impact on the Wi-Fi, they adopt a discontinuous transmission pattern such that transmissions comply to a given duty-cycle ($10$\% or $30$\% in our evaluation). In  LTE-LAA, a fixed duty-cycle of $50$\% and the same threshold-based CCA as in Wi-Fi is used, where we consider levels of $100$\% and $50$\% of the Wi-Fi threshold. 

The deployment of both systems follows the base system with an LTE base station density $\lambda = 0.001$, which is $10$\% of the \mbox{Wi-Fi} node density. The Wi-Fi nodes move at a pedestrian speed of $6$~km/h; the LTE base stations are static. A transmission duration of $20$ slots ($10$~ms) is used in both systems, which corresponds to LTE-LAA Class~4 access priority~\cite{Rev3_b}.

\begin{figure}[htbp]
 \centering
  \resizebox{.49\textwidth}{!}{
   \definecolor{mycolor1}{rgb}{0.39216,0.83137,0.07451}%
\begin{tikzpicture}

\begin{axis}[%
width=4.521in,
height=3.278in,
at={(0.1in,3.4in)},
xmin=0.5,
xmax=5,
ymin=-14,
ymax=-2,
ylabel style={font=\bfseries\color{white!15!black}},
ylabel={NMSE in dB},
axis background/.style={fill=white},
title style={font=\bfseries},
title={(a) Coexistence of Wi-Fi with LTE-U},
axis x line*=bottom,
axis y line*=left,
xmajorgrids,
ymajorgrids,
legend style={font=\scriptsize,at={(0.4,0.02)}, anchor=south west, legend cell align=left, align=left, draw=white!15!black}
]
\addplot [color=black, line width=1.0pt, mark=square, mark options={solid, black}]
  table[row sep=crcr]{%
0.5	-2.42248655191522\\
1	-2.41481678629166\\
1.5	-2.41520615062957\\
2	-2.42991944549109\\
2.5	-2.41951462805197\\
3	-2.42666214253987\\
3.5	-2.42258510198217\\
4	-2.39399850284857\\
4.5	-2.40715195612853\\
5	-2.39501429925667\\
};
\addlegendentry{Baseline MVP}

\addplot [color=blue, line width=1.0pt, mark=o, mark options={solid, blue}]
  table[row sep=crcr]{%
0.5	-12.4838786586338\\
1	-9.5968025147104\\
1.5	-7.99409664845781\\
2	-6.84440787993423\\
2.5	-6.04742242481903\\
3	-5.33629338210952\\
3.5	-4.84958687638759\\
4	-4.40216031848641\\
4.5	-4.02186395450911\\
5	-3.74986503929661\\
};
\addlegendentry{Wi-Fi no CCA}

\addplot [color=blue, line width=1.0pt]
  table[row sep=crcr]{%
0.5	-9.35788393524672\\
1	-7.39615033068475\\
1.5	-6.09673108587246\\
2	-5.15584905684293\\
2.5	-4.46979458134121\\
3	-3.89760951022443\\
3.5	-3.44458605638845\\
4	-3.06383484041846\\
4.5	-2.73628628402886\\
5	-2.43967166314341\\
};
\addlegendentry{Wi-Fi-CCA}

\addplot [color=red, dashed, line width=1.0pt]
  table[row sep=crcr]{%
0.5	-11.7742144743033\\
1	-8.85995850815468\\
1.5	-7.16609617620093\\
2	-6.08347011948308\\
2.5	-5.28893036774761\\
3	-4.57920853068432\\
3.5	-4.03875485379283\\
4	-3.64368107048498\\
4.5	-3.19625495771462\\
5	-2.89116644685025\\
};
\addlegendentry{Wi-Fi-CCA + LTE-U 10\% Duty Cycle}

\addplot [color=green, dashdotted, line width=1.0pt]
  table[row sep=crcr]{%
0.5	-13.1114393244184\\
1	-10.2264375552253\\
1.5	-8.61153947914889\\
2	-7.55529371606179\\
2.5	-6.68619983598515\\
3	-6.08617448954008\\
3.5	-5.56154016600225\\
4	-5.13373686211094\\
4.5	-4.79027024637714\\
5	-4.50020568611158\\
};
\addlegendentry{Wi-Fi-CCA + LTE-U 30\% Duty Cycle}

\end{axis}

\begin{axis}[%
width=4.521in,
height=3.278in,
at={(0.1in,0.1in)},
xmin=0.5,
xmax=5,
xlabel style={font=\bfseries\color{white!15!black}},
xlabel={Prediction horizon in ms},
ymin=-14,
ymax=-2,
ylabel style={font=\bfseries\color{white!15!black}},
ylabel={NMSE in dB},
axis background/.style={fill=white},
title style={font=\bfseries},
title={(b) Coexistence of Wi-Fi with LTE-LAA},
axis x line*=bottom,
axis y line*=left,
xmajorgrids,
ymajorgrids,
legend style={font=\scriptsize,at={(0.15,0.02)}, anchor=south west, legend cell align=left, align=left, draw=white!15!black}
]
\addplot [color=black, line width=1.0pt, mark=square, mark options={solid, black}]
  table[row sep=crcr]{%
0.5	-2.42248655191522\\
1	-2.41481678629166\\
1.5	-2.41520615062957\\
2	-2.42991944549109\\
2.5	-2.41951462805197\\
3	-2.42666214253987\\
3.5	-2.42258510198217\\
4	-2.39399850284857\\
4.5	-2.40715195612853\\
5	-2.39501429925667\\
};
\addlegendentry{Baseline MVP}

\addplot [color=blue, line width=1.0pt, mark=o, mark options={solid, blue}]
  table[row sep=crcr]{%
0.5	-12.4838786586338\\
1	-9.5968025147104\\
1.5	-7.99409664845781\\
2	-6.84440787993423\\
2.5	-6.04742242481903\\
3	-5.33629338210952\\
3.5	-4.84958687638759\\
4	-4.40216031848641\\
4.5	-4.02186395450911\\
5	-3.74986503929661\\
};
\addlegendentry{Wi-Fi no CCA}

\addplot [color=blue, line width=1.0pt]
  table[row sep=crcr]{%
0.5	-9.35788393524672\\
1	-7.39615033068475\\
1.5	-6.09673108587246\\
2	-5.15584905684293\\
2.5	-4.46979458134121\\
3	-3.89760951022443\\
3.5	-3.44458605638845\\
4	-3.06383484041846\\
4.5	-2.73628628402886\\
5	-2.43967166314341\\
};
\addlegendentry{Wi-Fi-CCA}

\addplot [color=red, dashed, line width=1.0pt]
  table[row sep=crcr]{%
0.5	-10.0312764337889\\
1	-7.82628482825919\\
1.5	-6.43191171946103\\
2	-5.50415017685571\\
2.5	-4.78437396283718\\
3	-4.21039971155989\\
3.5	-3.75345810995442\\
4	-3.36001575391402\\
4.5	-3.01668591603859\\
5	-2.74414728086582\\
};
\addlegendentry{Wi-Fi-CCA + LTE-LAA 50\% Duty Cycle (Wi-Fi thres)}

\addplot [color=mycolor1, dashdotted, line width=1.0pt]
  table[row sep=crcr]{%
0.5	-9.8067505625519\\
1	-7.64887568179555\\
1.5	-6.26598019444248\\
2	-5.31974929760826\\
2.5	-4.64800544852003\\
3	-4.06455220851957\\
3.5	-3.63206649798862\\
4	-3.21494913852276\\
4.5	-2.91919803952196\\
5	-2.62188744204913\\
};
\addlegendentry{Wi-Fi-CCA + LTE-LAA 50\% Duty Cycle (0.5.Wi-Fi thres)}

\end{axis}

\end{tikzpicture}%
 }
\caption{Predictor performance in a scenario with coexistence of \mbox{Wi-Fi} and LTE using (a) LTE-U or (b) LTE-LAA. The \mbox{Wi-Fi} interference coherence time for $\theta = 0.25$ is $5.3$~ms for a $6$~km/h node speed. Both technologies use a message length of $20$~slots, equivalent to $10$~ms. }
\label{Fig:Coexistence}
\end{figure}
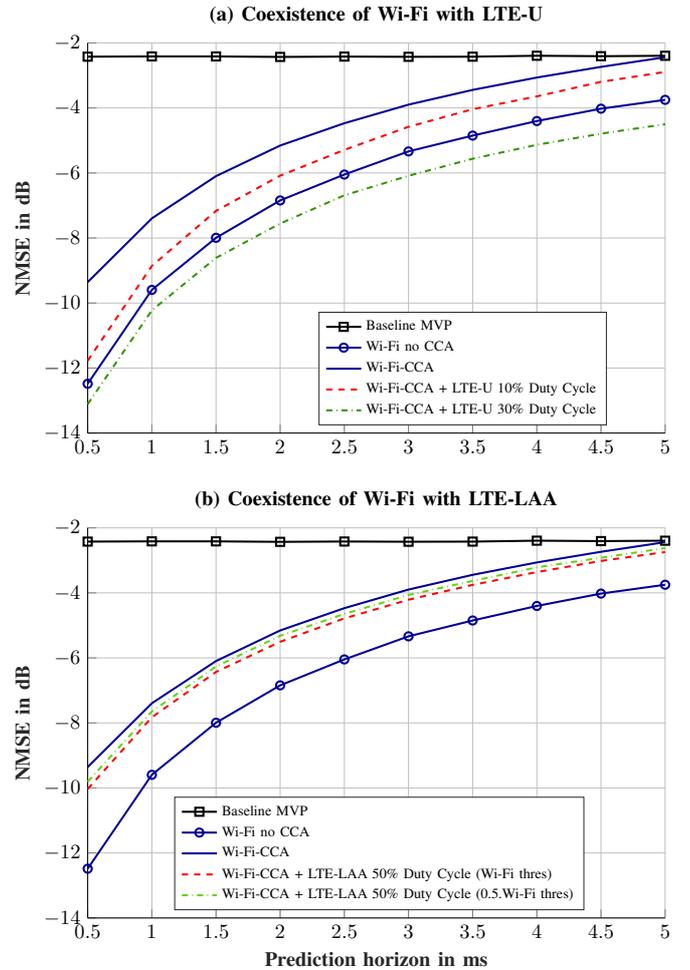  
\vspace{0.5cm}

\subsubsection{Performance Results}
Fig.~\ref{Fig:Coexistence} shows the results for the coexistence setups, mapping the prediction horizons to the signal timing. Since the predictor design does not account for the CCA, the performance for Wi-Fi without CCA (design assumption) is shown for reference. Subfigure~(a) shows the case of Wi-Fi and LTE-U coexistence. The CCA modifies the Wi-Fi interference dynamics, as can be seen from the degraded performance of Wi-Fi using CCA with respect to Wi-Fi without CCA (see also Fig.~\ref{Fig:Sensitivity_1}). The traces for Wi-Fi coexisting with LTE-U are below the trace for \mbox{Wi-Fi} using CCA alone, improving the predictor performance. This happens because the additional LTE-U interferers do not perform CCA and therefore make the aggregated interference dynamics closer to that for which the predictor is designed. For high duty-cycles, LTE-U increases the overall interference correlation making the predictor perform even better than for Wi-Fi without CCA. 
For the coexistence with LTE-LAA (Subfigure~(b)), the impact of LTE on the predictor performance is minimal.\footnote{Robustness against LTE interference does not necessarily imply fair coexistence between Wi-Fi and LTE-LAA. Prediction robustness in the context of this article indicates that the predictor is suitable for resource allocation applications, where it is used to find an optimal transmission slot \cite{schmidt2011prediction}.} Because  LTE-LAA  also uses a CCA mechanism, the dynamics of the overall interference is almost unchanged. Overall, we find that the predictor is robust to unknown interference of similar dynamics (LTE-LAA). Furthermore, the predictor sensitivity to the channel access method (CCA or not) is significant. The latter suggests that further models of channel access should be incorporated into the predictor~design.

\section{Conclusions and Outlook}
\label{SEC:Conclusions}

This article showed how to harness the stochastic characterization of interference dynamics for the task of interference prediction in wireless mobile networks. Our predictor designed as a recursive filter can be parameterized offline and shows reasonable computational complexity. Its performance analysis under matched and unmatched system conditions has demonstrated the prediction accuracy and robustness. Its versatility has been validated in multiple scenarios with parameters from LTE, sensor networks, and coexistence of Wi-Fi and~LTE.

Future work will address additional theoretical and practical aspects of interference prediction. First, the use of nonlinear filters for deriving a higher accuracy predictor and the incorporation of alternative channel access schemes in the interference model should be addressed. Second, 
the predictor needs to be implemented and tested in programmable radios. Third, the impact of interference prediction on the network dynamics should be studied: in contrast to channel prediction, the use of interference prediction by multiple nodes changes the stochastic features of the interference process that is being predicted, and its feedback effect needs to be accounted~for. 
 
\section*{Appendix A: Mapping $\rho(\tau)$ to an ARMA($p,q$) Model}
Starting from \eqref{eq;ARMAdef2_aux} and following \cite[Ch.~17]{Pollock}, we let $\tau$ take values from $0$ to $p+q$ to generate the following equation set 
\begin{equation}
	\begin{bmatrix}
	\rho(0)    &\cdots &\rho(p) \\ 
	\rho(1)    &\cdots &\rho(0) \\ 
	\vdots     &       &\vdots  \\ 
	\rho(q)    &\cdots &\rho(p\!-\!q) \\ 
	\vdots     &\ddots &\vdots  \\ 
	\rho(q\!+\!p)  &\cdots &\rho(q) 
	\end{bmatrix} 
	\begin{bmatrix}
	a_0\\ 
	a_1\\ 
	\vdots\\ 
	a_p
	\end{bmatrix}=
	\begin{bmatrix}
	b_0    &b_0    &\cdots &b_0    \\ 
	0      &b_0    &\cdots &b_0    \\ 
	\vdots &\vdots &\ddots &\vdots \\ 
	0      &0      &\cdots &b_0    \\ 
	\vdots &\vdots &       &\vdots \\ 
	0      &0      &\cdots &0 
	\end{bmatrix}
	\begin{bmatrix}
	b_0\\ 
	b_1\\ 
	\vdots\\ 
	b_q
	\end{bmatrix}.
	\label{eq;Param1}
\end{equation}
Specializing for $\tau=q+1,\ldots,q+p$ we have
\begin{equation}
\begin{bmatrix}
	\rho(q)       &\cdots       &\rho(q\!-\!p\!+\!1\!) \\ 
	\rho(q\!+\!1) &\cdots &\rho(q\!-\!p) \\ 
	\vdots    &\ddots &\vdots  \\ 
	\rho(q\!+\!p) &\cdots &\rho(q) 
	\end{bmatrix}\!\! 
	\begin{bmatrix}
	a_1\\ 
	a_2\\ 
	\vdots\\ 
	a_p
	\end{bmatrix}\!\!=\!\!-a_0\!
	\begin{bmatrix}
	\rho(q\!+\!1)\\ 
	\rho(q\!+\!2)\\ 
	\vdots\\ 
	\rho(q\!+\!p)
	\end{bmatrix},
	\label{eq;Param2}
\end{equation}
which imposing $a_0=1$ is recognized as the Yule Walker equations to solve for the AR coefficients $a_1,\ldots,a_p$ \cite{RegaliaLibro}. 

For determining the MA coefficients, consider \eqref{eq:bn_extraction1_aux} and \eqref{eq:bn_extraction2_aux} from introducing $\psi(t)$. All terms are known from the solution of \eqref{eq;Param2}. Thus, by running $\tau$ from $0$ to $q$, we can generate the required set of equations to solve for $b_1,\ldots,b_q$ and $\sigma_\epsilon^2$: 
\begin{eqnarray}
	\label{eq:MatrixMnum}
	\begin{bmatrix}
	\psi(0)\\ 
	\psi(1)\\ 
	\vdots\\ 
	\psi(q)
	\end{bmatrix}\!\!&=&\!\!\sigma_\epsilon^2
	\begin{bmatrix}
	b_0        &b_1    &\cdots  &b_{q\!-\!1} &b_q \\ 
	b_1        &b_1    &\cdots  &b_q         &0   \\ 
	\vdots     &\vdots &        &\vdots      &\vdots \\ 
	b_q        &0      &\cdots  &0           &0 
	\end{bmatrix} \!\!
	\begin{bmatrix}
	b_0\\ 
	b_1\\ 
	\vdots\\ 
	b_q
	\end{bmatrix}\\
		\!\!&=&\!\!\sigma_\epsilon^2
	\begin{bmatrix}
	b_0        &b_1    &\cdots  &b_{q\!-\!1} &b_q \\ 
	0        &b_0    &\cdots  &b_{q\!-\!2} &b_{q\!-\!1}   \\ 
	\vdots     &\vdots &        &\vdots      &\vdots \\ 
	0        &0      &\cdots  &0           &b_0 
	\end{bmatrix} \!\!
	\begin{bmatrix}
	b_0\\ 
	b_1\\ 
	\vdots\\ 
	b_q
	\end{bmatrix},
	\label{eq:MatrixPrime}
\end{eqnarray}
which can be written as $
	{\boldsymbol\psi} = \sigma_\epsilon^2{\cal M}^\#{\bf b}=\sigma_\epsilon^2{\cal M}{\bf b}$, 
where ${\boldsymbol\psi}$ and ${\bf b}$ are the vector representations for $\psi(t)$ and the MA coefficients, respectively, and ${\cal M}^\#$ and ${\cal M}$ are the matrices in \eqref{eq:MatrixMnum} and \eqref{eq:MatrixPrime}. The goal is to find the solution vector~$\bf b$ for the nonlinear system $\boldsymbol\psi - \sigma_\epsilon^2{\cal M}^\#{\bf b}=0$,
which is found iteratively. In particular, we use the procedure from Tunnicliffe-Wilson~\cite{Wilson1969}, where the $r$th approximation to the solution is computed from the $(r-1)$th instance as
\begin{equation*}
	\label{eq:Tunnicliffe-Wilson}
	{\bf b}_r = {\bf b}_{r-1}+\left\{\sigma_\epsilon^2\left({\cal M}^\#+{\cal M}\right)\right\}_{r-1}^{-1}\left(\boldsymbol\psi - \sigma_\epsilon^2{\cal M}^\#{\bf b}\right)_{r-1}.
\end{equation*}
As initialization we set $\sigma_\epsilon^2=1$ and $b_1=b_2=\ldots=b_q=0$. Upon convergence we normalize to have $b_0=1$.

\bibliographystyle{IEEEtran}

\end{document}